\pdfoutput=1
\documentclass{article}
\usepackage{iclr2027_conference,times}

\usepackage{amsmath,amsfonts,amssymb}
\usepackage[hidelinks]{hyperref}
\usepackage{graphicx}
\usepackage{booktabs}
\usepackage{multirow}
\usepackage{array}
\usepackage{xcolor}
\usepackage{flafter,placeins}
\graphicspath{{figures/}}
\title{Which State Should Prediction Read?\\ A Mechanistic Analysis of State--Prediction Separation}

\author{Ryan Kim \\
University of Illinois at Urbana-Champaign\\
\texttt{ryankim4@illinois.edu}\thanks{Code, configs and results: \url{https://github.com/RyanKim17920/mechanistic-sps}; weights: \url{https://huggingface.co/ryankim17920/mechanistic-sps}.}}

\def\dsTrMacro{44.89}

\def\dsSPSMacro{45.71}

\def\dsTTSixMacro{45.21}

\def\dsSeqSixMacro{45.60}

\def\dsSeqSixPredMacro{45.62}

\def\dsTrTwentyFourMacro{45.93}

\def\dsTTTwelveMacro{46.75}

\def\dsSeqTwelveMacro{47.62}

\def\dsDiffSeqSixTTSix{+0.39 [$-$0.13, +0.89]}
\def\dsDiffSeqSixTr{+0.72 [+0.19, +1.24]}
\def\dsDiffSPSTr{+0.82 [+0.31, +1.34]}
\def\dsDiffTTSixTr{+0.33 [$-$0.18, +0.83]}
\def\dsDiffSeqSixSPS{$-$0.11 [$-$0.61, +0.39]}
\def\dsDiffSeqTwelveTTTwelve{+0.87 [+0.35, +1.40]}
\def\dsDiffSeqTwelveTrTwentyFour{+1.69 [+1.05, +2.33]}
\def\dsDiffTTTwelveTrTwentyFour{+0.82 [+0.20, +1.44]}
\def\dsDiffSeqSixPredTr{+0.74 [+0.10, +1.38]}

\iclrfinalcopy
\begin{document}

\maketitle
\suppressfloats[t]

\begin{abstract}
Each hidden state in a decoder-only Transformer does two jobs: it stores context for later tokens and predicts the next token. To study each job separately, we split them into two towers with separate weights. Attention patterns show a division of labor: state heads record each position's previous token, and prediction heads match against that record. Ablating those state heads hurts retrieval more than other predictions, and restoring one head after corrupting the context recovers part of the lost signal. Linear probes show that state keeps previous-token information through its final layer, while prediction drops it in later layers as it builds the next-token distribution. Restricting prediction's reads and lesioning layers show that prediction depends most on later state layers and distant positions. We therefore design \emph{Sequential}, in which every prediction layer reads the final state. At equal non-embedding parameters and FLOPs, Sequential beats the Transformer at all four sizes we test (up to 327M non-embedding parameters), cutting validation loss by 0.042--0.079 nats/token and raising zero-shot accuracy. It also leads on out-of-domain text at 92M and 327M and holds its loss lead across 32K-token documents at 92M. Against State--Prediction Separation (SPS), which uses nearly twice the FLOPs per token, Sequential matches or beats it per token at small sizes and beats it at equal training FLOPs.
\end{abstract}

\section{Introduction}

In a decoder-only Transformer, each token's hidden state does two jobs at once: it stores context that later tokens read as keys and values (K/V), and it predicts the next token. Because one vector carries both, standard interpretability tools cannot easily tell which computation serves which role. State--Prediction Separation (SPS; \citealp{monea2026sps}) assigns the two jobs to separate token slots and improves over the Transformer at 1.6--1.95$\times$ the FLOPs per token, but its streams share weights, which still hides what each computes. Separating the jobs also lets prediction read from any state layer, leaving two questions: what does each stream compute, and which state should prediction read?

Encoder--decoder architectures and cross-layer K/V reuse also decouple building context from using it \citep{vaswani2017attention,sun2024yoco,wu2024layer,brandon2024cla}, and machine translation has long studied which encoder layers a decoder should read \citep{bapna2018training,he2018layerwise,domhan2018much,liu2021understanding}. To address both questions inside a decoder-only language model, we give state and prediction separate towers with separate weights and analyze each tower independently.

Previous-token and induction heads \citep{elhage2021mathematical,olsson2022context,wang2022interpretability} let us trace information flow, while activation patching, ablation \citep{meng2023locating,goldowskydill2023localizing,zhang2024best,heimersheim2024how,mcgrath2023hydra}, and linear probes \citep{alain2018understanding,belinkov2021probing,belrose2025eliciting} isolate each component's role. Separate weights let these tools assign each computation to one stream. We start with \emph{Two-tower}, the direct separate-weight counterpart of SPS. Because read restrictions and layer lesions show that prediction relies most on later state levels and distant positions, we introduce \emph{Sequential}, where every prediction layer reads the final state, and apply the same analyses across all models. We find:
\begin{enumerate}
\item \textbf{Stream specialization (\S\S\ref{sec:circuit}--\ref{sec:needs}).} State heads encode each position's previous token while prediction heads perform the matching. Ablating state's previous-token heads hurts retrieval selectively, and read restrictions show that prediction relies most on later, distant state.
\item \textbf{Sequential outperforms the Transformer at equal cost (\S\S\ref{sec:arch}--\ref{sec:beyond}).} Validation loss is lower at all four sizes and zero-shot accuracy is higher at every size (significantly at three). Sequential also leads on out-of-domain text at 92M and 327M, keeps its loss lead across 32K-token documents, wins at equal total parameters, and beats Two-tower and a YOCO-style cross-decoder. Starting prediction as a fresh stream matters: continuing state's residual stream, as in YOCO, costs 0.025 nats/token.
\item \textbf{Sequential outperforms SPS at equal training compute (\S\ref{sec:eff}).} Using 55--62\% of SPS's FLOPs per token, Sequential matches SPS at 92M and beats it at smaller sizes on the same token budget. Given SPS's training FLOPs, Sequential beats SPS at both 92M and 327M.
\end{enumerate}

\FloatBarrier
\section{A Controlled Setting for State and Prediction}
\label{sec:setup}

\subsection{Transformer and SPS}
\label{sec:background}
\begin{figure}[t]
\centering
\includegraphics[width=\linewidth]{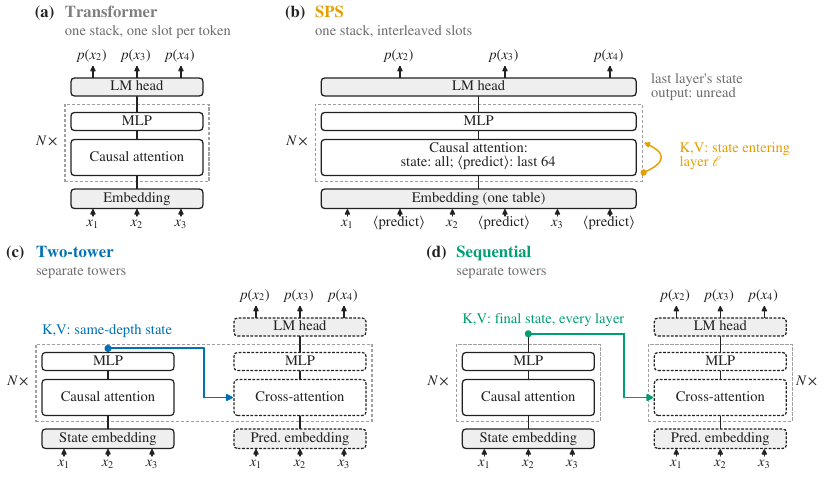}
\caption{Four ways to read state. (a) Transformer: one hidden state per token does both jobs. (b) SPS: interleaved state and prediction slots, shared weights. (c) Two-tower: prediction layer $\ell$ reads state level $\ell$. (d) Sequential: every prediction layer reads the final state. Dashed boxes: prediction towers; grey frames: repeated layers.}
\label{fig:teaser}
\end{figure}

The Transformer (Fig.~\ref{fig:teaser}a) is the single-stream baseline, with one slot per token and attention plus a multilayer perceptron (MLP) in each layer (12 layers at the 92M reference size). SPS (Fig.~\ref{fig:teaser}b) interleaves each token with a state slot and a prediction slot, both processed by the same weights, so $T$ tokens become $2T$ slots. Prediction slots receive a learned $\langle$predict$\rangle$ placeholder instead of the current token's embedding. Both slot types attend causally to all same-document state slots, but to prediction slots only within the preceding 64 token positions (the prediction-slot window). A \emph{state level} is a state layer's output, with level 0 the embedding. Prediction slots in layer $\ell$ attend to state as it enters that layer (level $\ell{-}1$).

\subsection{Two-tower and Sequential}
To expose the roles that SPS's shared weights hide, we give state and prediction separate weights in two \emph{separated models} (Tab.~\ref{tab:model-definitions}, Fig.~\ref{fig:teaser}c,d) with $a$ state and $b$ prediction layers.\footnote{Layers and heads are numbered from 1.} Two-tower, the direct counterpart of SPS with separate weights per stream, uses a \emph{same-depth read}: prediction layer $\ell$ reads state level $r(\ell)=\ell$. Sequential uses a \emph{final-state read}: $r(\ell)=a$ at every prediction layer. Sequential is the design our analysis motivates (\S\ref{sec:arch}); we define it here to analyze both reads. Each prediction layer has its own K/V projections of the state it reads, and prediction at $t$ reads only state at positions $\le t$ and never attends to other prediction positions. The last prediction layer's output is used for next-token prediction, which is where the cross-entropy loss is applied. We focus on depths $a{+}b$ of 6+6 and 12+12 because they match the 92M Transformer and SPS references: at 6+6 the separated models match the Transformer in non-embedding parameters and FLOPs, and at 12+12 they match SPS's FLOPs at twice the parameters. A size ladder covers other scales (\S\ref{sec:models}). Shared-weight versions and single-change controls: Apps.~\ref{app:sharing} and~\ref{app:full}.

\begin{table}[t]
\centering
\footnotesize
\caption{The main models differ in weights and in the state level prediction reads. Main depths: $a=b\in\{6,12\}$. Own embedding: the current token through a separate prediction embedding table.}
\label{tab:model-definitions}
\setlength{\tabcolsep}{2pt}
\begin{tabular*}{\linewidth}{@{\extracolsep{\fill}}llclcc@{}}
\toprule
Model & Weights & \shortstack{State level read by\\prediction layer $\ell$} & Prediction input & LM head & Seeds \\
\midrule
Transformer & -- & -- & -- & untied & 2 \\
SPS & shared & $\ell{-}1$ (input to layer $\ell$) & $\langle$predict$\rangle$ token & untied & 2 \\
Two-tower $a{+}b$ & separate & $\ell$ & own embedding & untied & 2 \\
Sequential $a{+}b$ & separate & final & own embedding & untied & 2 \\

\bottomrule
\end{tabular*}
\end{table}

\subsection{Training and Evaluation}
\label{sec:models}
All models train on 20B tokens of FineWeb-Edu, a filtered web-text corpus, with the GPT-2 tokenizer, width 768 (size S), 12 heads, context length 4096 and one learning-rate schedule, not tuned per model, that decays over the final 2B tokens (App.~\ref{app:training}). The size ladder (\S\ref{sec:eff}, Tab.~\ref{tab:scaling}, App.~\ref{app:ladder}) has sizes XXS, XS, S and M, with separated depths 3+3, 4+4, 6+6 and 12+12 (Transformer: $a{+}b$ layers), widths 384, 512, 768 and 1024, and 11.5M, 27.3M, 92M and 327M non-embedding parameters. We report negative log-likelihood (NLL) in nats/token, forward FLOPs/token, parameters, zero-shot accuracy on eight tasks, bits per byte on five out-of-domain corpora, and passkey and multi-key needle-in-a-haystack (NIAH) retrieval (App.~\ref{app:evalproto}). Results are means over two seeds unless marked, and $\pm$ is half the range, so mean $\pm$ spans both seeds. We do not call differences under 0.005 nats/token, about 2.4 pooled seed standard deviations (App.~\ref{app:evalproto}).

\subsection{Analysis Methods}
\label{sec:methods}
The analyses first separate previous-token encoding from matching, then test which state prediction needs (methods and control gates: App.~\ref{app:methods}). \emph{Previous-token mass} is the attention weight on the preceding token, and a \emph{previous-token head} attends mainly there. \emph{Induction lift}, attention to continuations of earlier occurrences of the current token divided by the uniform-attention baseline, locates matching \citep{olsson2022context,elhage2021mathematical}. \emph{Induction tokens} are targets that complete a bigram first seen more than 64 positions earlier, beyond SPS's prediction-slot window, so they could be copied from context.

To test whether prediction's retrieval depends on these heads, we ablate the three state heads with highest previous-token mass. We compare the rise in induction-token loss against low-mass control heads, and the drop in induction-token accuracy against random and pattern-matched control heads (heads with similar attention statistics), with Holm--Bonferroni correction \citep{holm1979simple}; a post hoc variant excludes layer-1 heads from the control pools (\emph{layer-1 exclusion}). \emph{Activation patching} tests whether restoring one clean state-head output after corrupting an earlier repeated segment recovers retrieval.

Read restrictions, applied at evaluation without retraining, limit prediction to the nearest $N$ state positions or to the first or last $m$ state levels; a prediction layer whose level is excluded reads the nearest kept level. \emph{Whole-layer lesions} zero a layer's residual contribution within a stream. Linear \emph{role probes} measure, by held-out NLL, the previous-token and next-token information readable at each layer (baselines: App.~\ref{app:probes}).

\FloatBarrier
\section{How the Separated Streams Divide Retrieval}
\label{sec:circuit}
The separated streams split the induction circuit: state records each position's previous token, and prediction matches against that record. Across all four separated models, state heads carry the bulk of previous-token attention (peak mass 0.71--0.89 versus 0.17--0.39 in prediction), whereas prediction heads achieve the higher mean induction lift (Tab.~\ref{tab:circuit}). The split is a tendency rather than a strict partition: Two-tower 12+12's state heads still reach a mean lift of 3.02. In SPS, state queries attend to the previous token primarily at its prediction slot (mass 0.80 versus 0.26 at its state slot). These circuit results, along with the probes and read-window costs below (\S\ref{sec:reach}), replicate on both seeds of 327M Sequential (App.~\ref{app:interp-m}).

\begin{table}[t]
\centering
\footnotesize
\caption{State heads carry the previous-token attention, prediction heads the higher induction lift. S/P: state/prediction queries; SPS rows split key slots. Prev.\ mass: the stream's highest per-head previous-token attention. $\Delta$NLL: three strongest previous-token state heads (Transformer: of its one stream) / low-mass controls ablated. Holm: induction-token accuracy drop against random / pattern-matched controls; all: significant with and without layer-1 exclusion; excl.: only with it; --: neither.}
\label{tab:circuit}
\setlength{\tabcolsep}{3pt}
\begin{tabular*}{\linewidth}{@{\extracolsep{\fill}}lcccccc}
\toprule
Model & Prev.\ mass & Mean lift & \multicolumn{2}{c}{$\Delta$NLL, abl./ctrl.} & \multicolumn{2}{c}{Holm-specific, s1/s2} \\
\cmidrule(lr){4-5}\cmidrule(lr){6-7}
 & S/P & S/P & induct.\ tok. & all tok. & random & pattern \\
\midrule
Transformer & 0.75/-- & 3.68/-- & 0.300/0.018 & 0.096/0.025 & --/all & all/all \\
Two-tower 12+12 & 0.85/0.37 & 3.02/4.38 & 0.095/0.015 & 0.032/0.015 & --/all & all/all \\
Sequential 12+12 & 0.71/0.17 & 2.12/4.73 & 0.137/0.010 & 0.046/0.009 & all/all & all/all \\
Two-tower 6+6 & 0.89/0.39 & 3.51/4.67 & 0.151/0.020 & 0.058/0.014 & excl./-- & excl./excl. \\
Sequential 6+6 & 0.82/0.21 & 3.17/5.17 & 0.202/0.007 & 0.108/0.018 & all/-- & excl./excl. \\
\addlinespace
SPS, state keys & 0.26/0.47 & 2.25/4.87 & \multirow{2}{*}{0.120/0.007} & \multirow{2}{*}{0.035/0.003} & \multirow{2}{*}{--/excl.} & \multirow{2}{*}{excl./excl.} \\
\quad pred.\ keys & 0.80/0.33 & 4.06/4.49 &  &  &  &  \\
\bottomrule
\end{tabular*}
\end{table}

Prediction's retrieval depends on this state record. Ablating the three strongest previous-token state heads raises induction-token loss by 0.095--0.202 nats/token in the separated models, compared to 0.007--0.020 for low-mass control heads (Tab.~\ref{tab:circuit}). This loss spike is selective: induction tokens degrade 1.9--3.0$\times$ as much as all tokens, versus 0.4--1.4$\times$ for controls. The corresponding accuracy drop passes every Holm-corrected test in Sequential 12+12 and a subset of tests in the other separated models, partly under the post hoc layer-1 exclusion (Tab.~\ref{tab:circuit}, Holm columns).

Single-head activation patching tests this link directly. Restoring the strongest previous-token state head recovers 0.23--0.50 of the corruption-induced logit change (answer versus distractor) in both seeds of both Sequential models and in one seed of each Two-tower model; the remaining Two-tower seeds recover 0.09 and 0.10, at the edge of the $\pm0.10$ random-donor band (Tab.~\ref{tab:patching}). Recovery is small in SPS and near zero when both towers share one weight stack (App.~\ref{app:sharing}), so separate weights are needed to isolate this circuit at the head level (App.~\ref{app:patching}).

\subsection{Previous-Token Information Reaches the Top State Layer}
\label{sec:roles}
\begin{figure}[t]
\centering
\includegraphics[width=\linewidth]{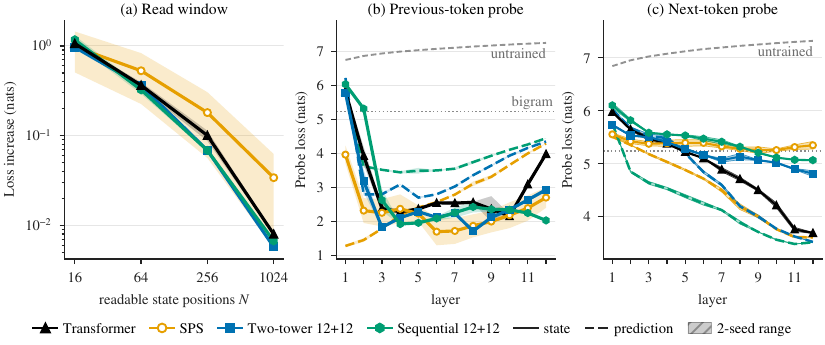}
\caption{Prediction relies on distant state, and at the top layer state holds the previous token while prediction holds the next. (a) Loss increase (nats/token, log scale) when prediction reads only the nearest $N$ state positions (Transformer: every layer restricted). (b) Previous-token and (c) next-token probe loss by layer; solid: state, dashed: prediction; Transformer (black): its single stream. Grey dashed: untrained; dotted: bigram. Seed means; shading: two-seed range.}
\label{fig:state-access}
\end{figure}

Linear probes show that state keeps its previous-token record through its top layer, where a final-state read still sees it. In SPS and the 12+12 separated models, state's previous-token probe loss stays low through the final layer (2.03 at layer 12 in Sequential 12+12; Fig.~\ref{fig:state-access}b; 6+6 models in App.~\ref{app:probes}). Prediction moves in the opposite direction: across every separated model and SPS, its previous-token probe loss climbs over later layers (Sequential 12+12: 3.44 at layer 4 to 4.45 at layer 12; Two-tower 12+12: 2.70 at layer 5 to 4.35; SPS: 1.28 at layer 1 to 4.29) while its next-token probe loss drops (Fig.~\ref{fig:state-access}c). The Transformer's single stream makes this handoff abruptly in its last two layers (2.16 at layer 10 to 3.98 at layer 12), whereas prediction transitions gradually. State itself never becomes a next-token predictor: its top-layer next-token probe loss (4.8--5.4 nats/token) stays near the bigram baseline (5.24), far above prediction's 3.5--3.7.

\section{What State Prediction Needs}
\label{sec:needs}
To determine which positions and layers prediction relies on, we restrict a trained model's reads at evaluation time and lesion whole layers.

\subsection{Prediction Needs Distant State}
\label{sec:reach}
Prediction cannot operate on a local window alone: restricting it to the nearest 64 state positions raises loss by 0.32--0.40 nats/token across the four separated models (Fig.~\ref{fig:state-access}a, Tab.~\ref{tab:readdepth_full}). Restricting every layer of the Transformer costs a similar 0.36 nats/token (App.~\ref{app:readreach}). The reliance on distant context is general, and in separated architectures all of it flows through state.

\begin{figure}[t]
\centering
\includegraphics[width=\linewidth]{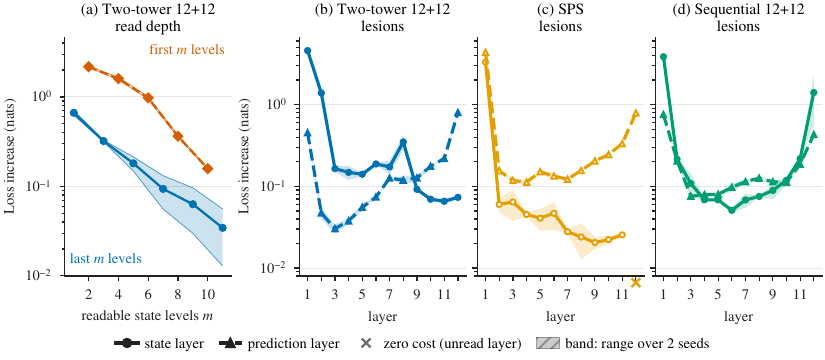}
\caption{Prediction depends more on later than on earlier state levels. Loss increase (nats/token, log scale). (a) Two-tower 12+12 keeping the last $m$ (solid) or first $m{+}1$ levels (dashed). (b--d) Lesion cost by layer in Two-tower 12+12, SPS and Sequential 12+12; circles: state, triangles: prediction.}
\label{fig:read-depth}
\end{figure}

\subsection{Later State Levels Matter More}
\label{sec:read}
Prediction also depends mainly on upper state layers. Keeping all positions and trained K/V projections in Two-tower 12+12, restricting prediction to only the first $m{+}1$ state levels costs 2.5--5.4$\times$ as much as keeping only the last $m$ levels (Fig.~\ref{fig:read-depth}a; App.~\ref{app:readdepth}). Because these restrictions are applied at evaluation time, they reflect what the trained model depends on rather than what a model trained on a single level would learn. When we instead let every prediction layer learn a mixture over all state levels during training (the \emph{multi-level read}, initialized on the final level; App.~\ref{app:multilevel}), the final level remains the single largest source in every layer, while later prediction layers draw roughly a third of their norm share from levels 1--9 (App.~\ref{app:mixweights}). This learned mixture builds on the final-state read and reduces loss further (\S\ref{sec:eff}).

Whole-layer lesions show that SPS makes progressively less use of later state layers and none of its top layer: its state-lesion cost drops roughly two-fold from early to late interior layers, and lesioning its top state layer costs nothing because nothing reads it (Fig.~\ref{fig:read-depth}c). Two-tower 12+12 uses both streams throughout, with state costs highest in early layers and prediction costs rising toward the top (Fig.~\ref{fig:read-depth}b; App.~\ref{app:lesions}). Early state layers are expensive to lesion because every subsequent level builds on them; read restrictions isolate which state level prediction actually reads.

\section{Final-State Reads Beat the Transformer and Two-tower}
\label{sec:arch}
\label{sec:final}
These findings point to one change. State keeps previous-token information to its top layer (\S\ref{sec:circuit}), and prediction depends most on later, distant state; even a learned mixture over all levels weights the final one highest (\S\ref{sec:needs}). Yet Two-tower exposes the final state only to its last prediction layer. \emph{Sequential} instead lets every prediction layer read the final state (Fig.~\ref{fig:teaser}d). Because swapping a trained Two-tower 6+6 to this read at evaluation raises loss by 0.68 nats/token (App.~\ref{app:readdepth}), we train all models from scratch.

Sequential outperforms the Transformer at matched non-embedding parameters and FLOPs at both budgets (Tab.~\ref{tab:2x2}). Sequential 6+6, which also matches the Transformer's K/V cache size (App.~\ref{app:wallclock}), finishes 0.064 nats/token below the 12-layer Transformer; at twice the parameters, Sequential 12+12 finishes 0.055 below a 24-layer Transformer. Changing only the read also beats Two-tower in both seeds, by 0.037 at 6+6 and 0.031 at 12+12. The analysis motivates reading later state, though not which tokens benefit most; \S\ref{sec:why} checks where the gain lands.

\begin{table}[t]
\centering
\caption{Sequential beats the Transformer and Two-tower at both budgets. Rows match their block's reference in non-embedding parameters and FLOPs unless the columns differ. Top-aligned: prediction layers read state levels 7--12. $\Delta$: NLL minus the reference's. Single-seed variants: App.~\ref{app:full}.}
\label{tab:2x2}
\scriptsize
\setlength{\tabcolsep}{2pt}
\begin{tabular*}{\linewidth}{@{\extracolsep{\fill}}lrrrlrc@{}}
\toprule
 & \multicolumn{2}{c}{Params (M)} & GFLOPs & NLL & & \\
\cmidrule(lr){2-3}
Model & non-emb. & total & /token & (nats/token) & $\Delta$ & Seeds \\
\midrule
\multicolumn{7}{@{}l}{\emph{The Transformer's non-embedding parameters and FLOPs; reference Sequential 6+6 (App.~\ref{app:ratio}, \ref{app:yoco})}} \\
Sequential 6+6 & 92.0 & 207.9 & 0.414 & 2.801$_{\pm.002}$ & -- & 2 \\
Two-tower 6+6 & 92.0 & 207.9 & 0.414 & 2.838$_{\pm<.001}$ & $+0.037$ & 2 \\
Sequential 3+9 / 9+3 & 92.0 & 207.9 & 0.414 & 2.813 / 2.815 & $+0.012$ / $+0.013$ & 1 \\
Residual continuation & 92.0 & 169.3 & 0.414 & 2.826 & $+0.025$ & 1 \\
Shared K/V, width-restored & 92.3 & 208.2 & 0.415 & 2.816 & $+0.015$ & 1 \\
YOCO-style cross-decoder & 86.1 & 163.4 & 0.402 & 2.850 & $+0.049$ & 1 \\
Transformer & 92.0 & 169.3 & 0.414 & 2.865$_{\pm.003}$ & $+0.064$ & 2 \\
SPS (1.83$\times$ FLOPs) & 92.0 & 169.3 & 0.756 & 2.797$_{\pm.001}$ & $-0.0048$ & 2 \\
\midrule
\multicolumn{7}{@{}l}{\emph{Twice the non-embedding parameters; reference Sequential 12+12 (App.~\ref{app:depthcap}, \ref{app:tf24})}} \\
Sequential 12+12 & 184.1 & 300.0 & 0.751 & 2.702$_{\pm<.001}$ & -- & 2 \\
Two-tower 12+12 & 184.1 & 300.0 & 0.751 & 2.733$_{\pm.001}$ & $+0.031$ & 2 \\
Two-tower 12+12, MLP moved to prediction & 184.1 & 300.0 & 0.751 & 2.737 & $+0.035$ & 1 \\
Two-tower 12+6, top-aligned, widened & 184.1 & 300.0 & 0.674 & 2.722 & $+0.021$ & 1 \\
Two-tower 12+6, top-aligned, widened more & 222.1 & 338.0 & 0.751 & 2.699 & $-0.003$ & 1 \\
Transformer 24L & 184.1 & 261.3 & 0.751 & 2.756 & $+0.055$ & 1 \\
\bottomrule
\end{tabular*}
\end{table}

Single-seed controls rule out three simpler explanations for the gain (Tab.~\ref{tab:2x2}, App.~\ref{app:archctrl}). First, extra \emph{prediction capacity} does not help: reallocating MLP width from state to prediction leaves Two-tower 12+12 at 2.737 (versus 2.733), and shifting layers between Sequential's towers raises loss from 2.801 (6+6) to 2.813 (3+9) and 2.815 (9+3). Second, \emph{prediction depth} matters less than reading later state: a top-aligned Two-tower 12+6 whose six prediction layers read state levels 7--12 beats Two-tower 12+12 when widened to match its non-embedding parameters (2.722 versus 2.733), and ties Sequential 12+12 when widened to match its FLOPs at 21\% more parameters (2.699 versus 2.702). Third, \emph{doubling parameters} alone does not explain the 12+12 lead: a 24-layer Transformer with the same non-embedding parameters and FLOPs reaches 2.756, trailing both Two-tower 12+12 (2.733) and Sequential 12+12 (2.702).

Sequential is structurally close to the cross-decoder of You Only Cache Once (YOCO; \citealp{sun2024yoco}), which also reads the final state at every upper layer, but YOCO continues the lower layers' residual stream and shares a single K/V projection across upper layers. Both choices hurt loss at equal FLOPs (Tab.~\ref{tab:2x2}, App.~\ref{app:yoco}): continuing state's residual alone costs 0.025 nats/token (whereas replacing Sequential's prediction embedding table with a single $\langle$predict$\rangle$ token barely changes loss, to 2.805), sharing one K/V set costs 0.015 even with width restored, and combining both in a YOCO-style model (6.4\% fewer non-embedding parameters, 3\% fewer FLOPs) finishes 0.049 behind Sequential 6+6. YOCO makes both choices to shrink the K/V cache; at equal FLOPs, starting prediction fresh and giving each layer its own K/V projections yields lower loss.

\subsection{Where the Gain Comes From}
\label{sec:why}
Sequential's gain over Two-tower is spread across token types rather than concentrated on copyable tokens: relative gains are roughly 1--2\% both on tokens appearing for the first time in a document and at every repeat distance (Tab.~\ref{tab:finalread}). Induction tokens account for 19.3\% of all tokens but only 8--14\% of the loss reduction (App.~\ref{app:finalread}). So while the copying circuit (\S\ref{sec:circuit}) exposed prediction's reliance on late, distant state, reading that state helps all token types.

Probes and lesions show how this shifts computation in the prediction tower. Because Sequential's prediction stream reads the final state from layer 1 onward, its next-token probe reaches 5.0 and 4.5 nats/token three to four layers earlier than in Two-tower 12+12 (Fig.~\ref{fig:state-access}c). Lesion costs also balance across both streams, making the top state layer the costliest state layer after layer 1 (Fig.~\ref{fig:read-depth}d, App.~\ref{app:lesions}).

\section{Scaling with Model Size and Compute}
\label{sec:eff}
We next test whether Sequential's advantage holds across model scales and against SPS, which spends 1.6--1.95$\times$ the Transformer's FLOPs per token.

\begin{figure}[!tb]
\centering
\includegraphics[width=\linewidth]{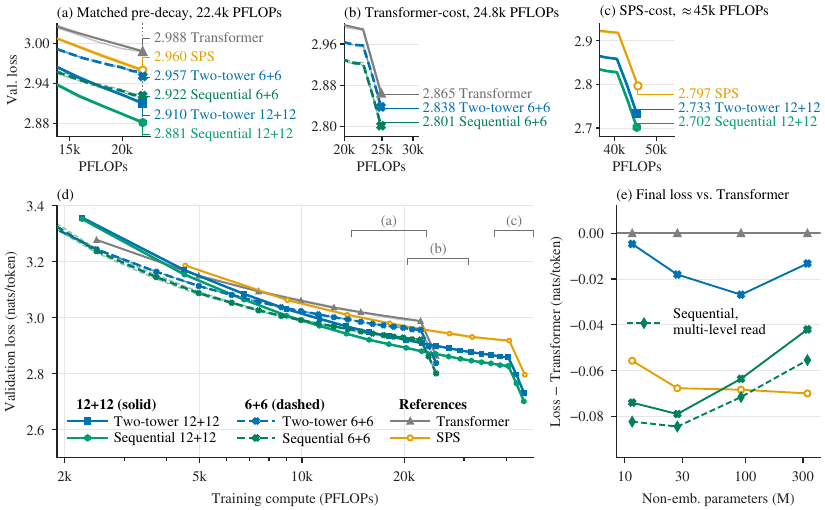}
\caption{Sequential 6+6 beats the Transformer at equal compute and nearly matches SPS with 55\% of its compute. Validation loss (nats/token) against training compute (PFLOPs, log scale); thin lines: single seeds. (a) Before learning-rate decay; (b, c) zoom on the ends of the runs that finish at Transformer and at SPS compute; (d) all runs in full. (e) Final loss minus the Transformer's by size; dashed: Sequential with a learned multi-level read (\S\ref{sec:read}).}
\label{fig:frontier}
\end{figure}

\paragraph{Across sizes.} Sequential beats the Transformer at all four sizes with a Transformer baseline, from 11.5M to 327M non-embedding parameters (Fig.~\ref{fig:frontier}e, Tab.~\ref{tab:scaling}, App.~\ref{app:ladder}). The extra prediction embedding table does not drive the gain: replacing it with a single $\langle$predict$\rangle$ token matches the Transformer's total parameter count while keeping a lead of 0.060 at 92M and 0.044 at 327M (App.~\ref{app:predinput}). At 767M we trained only Sequential (2.503); for rough context, under a different training protocol \citet{monea2026sps} report 2.537 for a Transformer and 2.470 for SPS of that shape, and the multi-level read lowers Sequential to 2.489 (App.~\ref{app:multilevel}).

\paragraph{Against SPS at equal tokens.} Using 55--62\% of SPS's FLOPs per token, Sequential beats SPS on the same token budget at 11.5M and 27.3M and ties it at 92M (2.801 versus 2.797, 0.0048 apart), recovering 93\% of SPS's gain over the Transformer at 55\% of its training compute (Fig.~\ref{fig:frontier}b,c). At 327M, SPS leads per token by 0.028. Before learning-rate decay, Sequential 6+6 lies 0.038 below SPS at equal compute (Fig.~\ref{fig:frontier}a). With tied heads across all three models, Sequential 6+6 finishes 0.010 above tied-head SPS and recovers 72--82\% of its gain (App.~\ref{app:tiedgroup}).

\paragraph{Against SPS at equal training FLOPs.} Given SPS's training FLOP budget at equal non-embedding parameters, Sequential beats SPS at both scales tested: Sequential 6+6 reaches 2.760 after 36.5B tokens versus SPS's 2.797, and 327M Sequential reaches 2.557 after 38.5B tokens versus 2.583 (0.025 before rounding; Sequential: one seed, SPS: two-seed means). Because the cheaper model trains longer at fixed compute, these runs see 1.83$\times$ and 1.92$\times$ as many unique tokens as SPS. At equal tokens before decay they stay within 0.004 of the 20B runs, bounding any effect of the extended data file and sampler well below the 0.025--0.037 margins (App.~\ref{app:isoflop}). Per forward FLOP, a Sequential model scaled up to each SPS model's FLOPs/token (interpolated along our size ladder, hence with more parameters) sits 0.19, 0.16, 0.11 and 0.07 nats/token below SPS from 11.5M to 327M (Fig.~\ref{fig:fwdfrontier}).

\paragraph{Wall-clock.} The FLOP savings carry over to training time: on the same block-sparse kernel and node, SPS takes 1.75$\times$ Sequential 6+6's step time, while Sequential 6+6 matches the Transformer's K/V cache size and step time (App.~\ref{app:wallclock}).

\paragraph{Margin by size.} Sequential's margin over the Transformer narrows at the largest size: 0.074, 0.079, 0.064 and 0.042 nats/token from 11.5M to 327M. App.~\ref{app:narrow} isolates two contributing factors: splitting depth across two towers and reading from a single state level.

\section{Beyond Validation Loss}
\label{sec:beyond}
Because validation loss is in-domain, we also evaluate zero-shot downstream tasks, out-of-domain corpora and long-context behavior (protocols: App.~\ref{app:evalproto}; full results: App.~\ref{app:beyond}).

\begin{table}[t]
\centering
\footnotesize
\caption{Beyond validation loss, Sequential improves on its block's Transformer. Model minus the block's Transformer: macro zero-shot accuracy (points, paired 95\% interval; $^{*}$: excludes zero), mean bits per byte over five out-of-domain corpora, and NLL at 16K--32K tokens after 32K continued pretraining (1.05B tokens at 32K context; \S\ref{sec:beyond}, Long context). $^1$: one seed; $^\dagger$: against the 12-layer Transformer. SPS's out-of-domain mean is above the Transformer's only because of code (1.52 versus 1.42 bits/byte); it is lower on the other four corpora.}
\label{tab:beyond}
\scriptsize
\setlength{\tabcolsep}{2pt}
\begin{tabular*}{\linewidth}{@{\extracolsep{\fill}}lrclccc@{}}
\toprule
 & Non-emb. & Macro & Accuracy minus Transf. & OOD bits/byte & \multicolumn{2}{c}{32K NLL minus Transf.\ (one seed)} \\
\cmidrule(lr){6-7}
Model & params (M) & acc.\ (\%) & (points, 95\% CI) & minus Transf. & PG19 & FW-Edu long \\
\midrule
Transformer & 92.0 & \dsTrMacro & -- & -- & -- & -- \\
SPS (1.83$\times$ FLOPs) & 92.0 & \dsSPSMacro & \dsDiffSPSTr$^{*}$ & $+0.005$ & $-0.058$ & $-0.090$ \\
Two-tower 6+6 & 92.0 & \dsTTSixMacro & \dsDiffTTSixTr & $-0.014$ & $-0.009$ & $-0.026$ \\
Sequential 6+6 & 92.0 & \dsSeqSixMacro & \dsDiffSeqSixTr$^{*}$ & $-0.029$ & $-0.043$ & $-0.076$ \\
Seq.\ 6+6, $\langle$predict$\rangle$ input$^1$ & 92.0 & \dsSeqSixPredMacro & \dsDiffSeqSixPredTr$^{*}$ & $-0.026$ & -- & -- \\
\midrule
Transformer 24L$^1$ & 184.1 & \dsTrTwentyFourMacro & -- & -- & -- & -- \\
Two-tower 12+12 & 184.1 & \dsTTTwelveMacro & \dsDiffTTTwelveTrTwentyFour$^{*}$ & $-0.022$ & $-0.106^{\dagger}$ & $-0.158^{\dagger}$ \\
Sequential 12+12 & 184.1 & \dsSeqTwelveMacro & \dsDiffSeqTwelveTrTwentyFour$^{*}$ & $-0.035$ & $-0.137^{\dagger}$ & $-0.205^{\dagger}$ \\
\bottomrule
\end{tabular*}
\end{table}

\textbf{Downstream tasks.} The validation-loss gains carry over to zero-shot benchmarks (Tab.~\ref{tab:beyond}). At equal non-embedding parameters and FLOPs, Sequential 6+6 beats the Transformer by \dsDiffSeqSixTr\ macro accuracy points (paired 95\% example-bootstrap interval, which omits seed noise and excludes zero in seed 2 but not seed 1), and Sequential 12+12 beats the 24-layer Transformer by \dsDiffSeqTwelveTrTwentyFour. Across the size ladder, Sequential's macro accuracy exceeds the Transformer's at every size and significantly at 27M, 92M and 327M (App.~\ref{app:downstream}). In head-to-head comparisons, Sequential 12+12 beats Two-tower 12+12 by \dsDiffSeqTwelveTTTwelve, while the 6+6 difference (\dsDiffSeqSixTTSix) is not significant. Sequential 6+6 ties SPS (\dsDiffSeqSixSPS) at 55\% of its FLOPs, though SPS leads on gold-answer bits per byte and by 0.96 accuracy points at 327M.

\textbf{Out-of-domain text and retrieval within 4K.} On five out-of-domain corpora, Sequential has lower bits per byte than the Transformer at 92M and 327M, and beats Two-tower on all but one tie (Tab.~\ref{tab:beyond}, App.~\ref{app:ood}); SPS leads on most sets at 327M. Within the 4K training window, Sequential 6+6 retrieves both passkeys (0.71 versus 0.59) and multi-key needles (0.54 versus 0.41) more accurately than the Transformer, though all models fall near chance once needles lie more than roughly 2.5K tokens back (App.~\ref{app:retrieval}).

\begin{figure}[!tb]
\centering
\includegraphics[width=\linewidth]{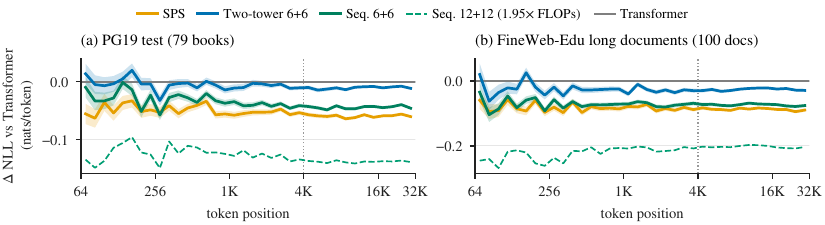}
\caption{After continued pretraining at 32K context (1.05B tokens; \S\ref{sec:beyond}), Sequential 6+6 keeps a steady lead over the Transformer. NLL minus the 12-layer Transformer's (nats/token) by position on (a) PG19 and (b) long FineWeb-Edu; dashed: Sequential 12+12, against the same Transformer; bands: 95\% intervals; dotted: 4K training length.}
\label{fig:longctx}
\end{figure}

\textbf{Long context.} After continuing all width-768 main models for 1.05B tokens at 32K context on PG19 and long FineWeb-Edu documents under one shared recipe (App.~\ref{app:evalproto}; one seed each), Sequential 6+6 maintains a steady loss lead across the full window: 0.043 nats/token below the Transformer on PG19 at 16K--32K (matching its 2K--4K margin) and 0.07--0.08 below on long FineWeb-Edu (Fig.~\ref{fig:longctx}, Tab.~\ref{tab:beyond}). SPS, at 1.95$\times$ the FLOPs at 32K, is 0.015 lower still. Long-range passkey retrieval does not follow the loss advantage: Sequential 6+6 beats the Transformer at 2K and 4K but trails at 16K (0.36 versus 0.55) and 32K (0.15 versus 0.28). Both 12+12 models retrieve best from 8K onward.

\section{Related Work}
\label{sec:related}
\paragraph{Extra computation for prediction.} Almost-Free SPS (AF-SPS, \citealp{langford2026free}; App.~\ref{app:afsps}) shares an MLP between SPS's streams, Decode-Branch Transformers (DBT, \citealp{liu2026decode}) move prediction off prefill, and parallel and looped designs add per-token compute with shared weights \citep{wu2025efficient,wu2025parallel,chen2025parallel,liu2026hidden}. We instead fix the cost and vary the read.

\paragraph{Final-state readers.} Encoder--decoder models read the encoder's final output \citep{wang2022what,elfeki2025return,zhang2025encoder,ho2024block}, and Layer-Condensed KV Cache (LCKV), You Only Cache Once (YOCO), Cross-Layer Attention and related schemes reuse keys and values across layers of a single stream to save memory \citep{wu2024layer,sun2024yoco,brandon2024cla,wu2025systematic,qiao2024swiftkv,gemma2025gemma3n}. Unlike these methods, which alter the full architecture, encode a separate source sequence, or continue a single residual stream, Sequential reads the final state of a dedicated state tower with separate weights (YOCO: \S\ref{sec:arch}; extended discussion: App.~\ref{app:relatedext}).

\section{Conclusion}
Separating state and prediction into distinct weight stacks isolates what each stream computes: state keeps previous-token context up to its top layer, and prediction matches against distant, late-layer state. Sequential gives every prediction layer the final state and beats the Transformer at equal non-embedding parameters and FLOPs at all four sizes tested, as well as SPS at equal training FLOPs. Open questions include how stream roles, tower depth splits, and the read interface behave at larger scales, and whether a learned multi-level read keeps Sequential's margin.

\paragraph{Limitations.}
Our full comparison covers 11.5M to 327M non-embedding parameters on 20B tokens of a single corpus (FineWeb-Edu) with a shared learning-rate schedule not tuned per model, and several controls, iso-FLOP runs, and 32K continuations use a single seed. Per token, SPS leads at 327M (\S\ref{sec:eff}), and Sequential 6+6 trails the Transformer on passkey retrieval at 16K and 32K (\S\ref{sec:beyond}). The untied-head SPS comparison was added after initial results and is post hoc. Training prediction towers on individual fixed state levels, evaluating a 6-layer SPS matched to Sequential 6+6's FLOPs, and testing LongBench and RULER remain for future work.

\bibliography{main}
\bibliographystyle{iclr2027_conference}

\appendix

The six parts follow the main text: setup (App.~\ref{app:setup}), mechanistic analyses for \S\S\ref{sec:circuit}--\ref{sec:needs} (App.~\ref{app:mech}), architecture controls for \S\ref{sec:arch} (App.~\ref{app:archctrl}), efficiency and scaling for \S\ref{sec:eff} (App.~\ref{app:scaling}), evaluations beyond validation loss for \S\ref{sec:beyond} (App.~\ref{app:beyond}) and related work (App.~\ref{app:related-top}).

\section{Setup and Reproducibility}
\label{app:setup}
How each model was built, trained and evaluated, completing \S\ref{sec:setup}.

\subsection{Models and Parameter and FLOP Accounting}
\label{app:full}
Let $a$ and $b$ be the numbers of state and prediction layers, $x_t$ the token, and $s^0_t=E_s(x_t)$, $p^0_t=E_p(x_t)$ its separate state and prediction embeddings. State is a stack of pre-norm causal blocks, and the read map is $r(\ell)=\ell$ for Two-tower and $r(\ell)=a$ for Sequential. With learned projections $W_Q,W_{KV},W_O$, scaled dot-product attention $\mathrm{Attn}$ and normalization inputs $P,R,M$ (prediction, read, MLP), prediction updates are
\begin{align*}
 q^\ell_t &= W^\ell_Q\,\mathrm{RMSNorm}^\ell_P(p^{\ell-1}_t), &
 (k^\ell_j,v^\ell_j) &= W^\ell_{KV}\,\mathrm{RMSNorm}^\ell_R(s^{r(\ell)}_j),\\
 \widetilde p^\ell_t &= p^{\ell-1}_t+W^\ell_O\,\mathrm{Attn}(q^\ell_t,k^\ell_{j\le t},v^\ell_{j\le t}), &
 p^\ell_t &= \widetilde p^\ell_t+\mathrm{MLP}^\ell(\mathrm{RMSNorm}^\ell_M(\widetilde p^\ell_t)).
\end{align*}
Queries and keys receive rotary position embeddings, and the attention mask excludes other documents. Each prediction layer has its own read normalization and K/V projection, and nothing from prediction is written back to state. A final normalization and LM head predict $x_{t+1}$ from $p^b_t$.

Each control changes one aspect (details in the section cited):
\begin{itemize}
\item MLP realloc.: Two-tower 12+12 with state/prediction MLP widths 1152/3456 at unchanged cost (App.~\ref{app:depthcap}).
\item $\langle$predict$\rangle$ input: a learned $\langle$predict$\rangle$ token replaces Sequential's prediction embedding table; tying also reuses the state embedding as LM head (App.~\ref{app:predinput}).
\item Top-aligned Two-tower 12+6 reads levels 7--12; its widened versions use prediction MLP width 5824 (parameter-matched) or 8576 (FLOP-matched) (App.~\ref{app:depthcap}). Sequential 3+9 and 9+3 vary the depth split (App.~\ref{app:ratio}).
\item Residual continuation starts prediction from the state's final residual; shared K/V computes one K/V set from the final state for all prediction layers; together they give a YOCO-style cross-decoder (App.~\ref{app:yoco}).
\item Multi-level read: each prediction layer reads a learned mixture $\sum_{l=1}^{a} w_{\ell l}\,s^l$, initialized one-hot on $l=a$ (App.~\ref{app:multilevel}).
\item Tied-head Transformer/SPS reuse their input embeddings as LM heads; Shared Two-tower/Sequential 12+12 reuse one stack (App.~\ref{app:sharing}); AF-SPS also shares a gated MLP (App.~\ref{app:afsps}).
\end{itemize}

Non-embedding parameters exclude the vocabulary tables (input embedding, prediction embedding where present, untied LM head; each $50304\times d$). Forward FLOPs/token are computed analytically at $T=4096$, counting attention scores over visible keys, and training compute is $3\times$ forward FLOPs/token $\times$ tokens.

\begin{table}[t]
\centering
\scriptsize
\caption{Cost and final validation NLL of every width-768 arm. Read: state level read by prediction layer $\ell$ ($\ell{-}1$: state entering layer $\ell$; mix: learned mixture; one K/V: one set for all prediction layers). NLL: seed 1 / seed 2; ``A / B'' rows: two splits, one seed each. $^{*}$: seeds 2 and 3. $^{\dagger}$: 36.5B tokens (App.~\ref{app:isoflop}).}
\label{tab:roster_full}
\setlength{\tabcolsep}{2pt}
\begin{tabular*}{\linewidth}{@{\extracolsep{\fill}}lllrrrll@{}}
\toprule
Model & Head / pred.\ input & Read & \multicolumn{2}{c}{Params (M)} & GFLOPs & NLL per seed & Mean \\
\cmidrule(lr){4-5}
 & & & total & non-emb. & /token & (nats/token) & \\
\midrule
\multicolumn{8}{@{}l}{\emph{Single stream and SPS}} \\
Transformer & untied & -- & 169.3 & 92.0 & 0.414 & 2.862 / 2.868 & 2.865 \\
Transformer & tied & -- & 130.7 & 92.0 & 0.414 & 2.851 & -- \\
Transformer 24L & untied & -- & 261.3 & 184.1 & 0.751 & 2.756 & -- \\
SPS & untied / $\langle$predict$\rangle$ & $\ell{-}1$ & 169.3 & 92.0 & 0.756 & 2.798 / 2.795 & 2.797 \\
SPS$^{*}$ & tied / $\langle$predict$\rangle$ & $\ell{-}1$ & 130.7 & 92.0 & 0.756 & 2.806 / 2.809 & 2.807 \\
AF-SPS & tied / $\langle$predict$\rangle$ & $\ell{-}1$ & 130.7 & 92.1 & 0.595 & 2.826 & -- \\
\midrule
\multicolumn{8}{@{}l}{\emph{Separate weights, 6+6 and other 12-layer splits}} \\
Two-tower 6+6 & untied / own table & $\ell$ & 207.9 & 92.0 & 0.414 & 2.838 / 2.838 & 2.838 \\
Sequential 6+6 & untied / own table & final & 207.9 & 92.0 & 0.414 & 2.803 / 2.800 & 2.801 \\
Sequential 6+6 & untied / $\langle$predict$\rangle$ & final & 169.3 & 92.0 & 0.414 & 2.805 & -- \\
Sequential 6+6 & tied / $\langle$predict$\rangle$ & final & 130.7 & 92.0 & 0.414 & 2.815 / 2.819 & 2.817 \\
Sequential 6+6, multi-level read & untied / own table & mix & 207.9 & 92.0 & 0.414 & 2.793 & -- \\
Sequential 3+9 / 4+8 & untied / own table & final & 207.9 & 92.0 & 0.414 & 2.813 / 2.808 & -- \\
Sequential 8+4 / 9+3 & untied / own table & final & 207.9 & 92.0 & 0.414 & 2.804 / 2.815 & -- \\
\midrule
\multicolumn{8}{@{}l}{\emph{YOCO decomposition (App.~\ref{app:yoco})}} \\
Residual continuation 6+6 & untied / none & final & 169.3 & 92.0 & 0.414 & 2.826 & -- \\
Residual continuation 6+6, detached & untied / none & final & 169.3 & 92.0 & 0.414 & 2.846 & -- \\
Shared K/V 6+6, width-restored & untied / own table & final, one K/V & 208.2 & 92.3 & 0.415 & 2.816 & -- \\
YOCO-style cross-decoder 6+6 & untied / none & final, one K/V & 163.4 & 86.1 & 0.402 & 2.850 & -- \\
\midrule
\multicolumn{8}{@{}l}{\emph{Separate weights, 24 layers}} \\
Two-tower 12+12 & untied / own table & $\ell$ & 300.0 & 184.1 & 0.751 & 2.732 / 2.733 & 2.733 \\
Two-tower 12+12, MLP realloc. & untied / own table & $\ell$ & 300.0 & 184.1 & 0.751 & 2.737 & -- \\
Two-tower 12+12, shared attn. & untied / own table & $\ell$ & 271.7 & 155.8 & 0.751 & 2.767 & -- \\
Two-tower 12+6, top-aligned & untied / own table & $\ell{+}6$ & 251.3 & 135.4 & 0.577 & 2.757 & -- \\
Two-tower 12+6, widened (param.) & untied / own table & $\ell{+}6$ & 300.0 & 184.1 & 0.674 & 2.722 & -- \\
Two-tower 12+6, widened (FLOPs) & untied / own table & $\ell{+}6$ & 338.0 & 222.1 & 0.751 & 2.699 & -- \\
Sequential 12+12 & untied / own table & final & 300.0 & 184.1 & 0.751 & 2.702 / 2.701 & 2.702 \\
Sequential 12+12, multi-level read & untied / own table & mix & 300.0 & 184.1 & 0.751 & 2.690 & -- \\
Sequential 12+12, input-dep.\ read & untied / own table & attn.\ mix & 300.0 & 184.1 & 0.751 & 2.694 & -- \\
\midrule
\multicolumn{8}{@{}l}{\emph{One shared stack run twice}} \\
Shared Two-tower 12+12 & tied / $\langle$predict$\rangle$ & $\ell{-}1$ & 130.7 & 92.0 & 0.722 & 2.814 & -- \\
Shared Sequential 12+12 & tied / $\langle$predict$\rangle$ & final, one K/V & 131.8 & 93.2 & 0.725 & 2.827 / 2.826 & 2.826 \\
Shared Sequential 12+12 & untied / own table & final & 207.9 & 92.0 & 0.751 & 2.806 & -- \\
\midrule
\multicolumn{8}{@{}l}{\emph{Trained to SPS's FLOPs (App.~\ref{app:isoflop})}} \\
Sequential 6+6, 36.5B tokens$^{\dagger}$ & untied / own table & final & 207.9 & 92.0 & 0.414 & 2.760 & -- \\
\bottomrule
\end{tabular*}
\end{table}

\subsection{Training Recipe and Protocol by Run Family}
\label{app:training}
\label{app:issues}
All runs share the base setup below (completing \S\ref{sec:models}), with per-family differences listed in Tab.~\ref{tab:protocol}. The Transformer uses a 2304-unit SwiGLU MLP, RoPE \citep{su2023roformer}, RMSNorm, and 38.6M input-embedding parameters. Models train on FineWeb-Edu 100BT \citep{penedo2024fineweb} with the GPT-2 tokenizer \citep{radford2019language} using AdamW \citep{loshchilov2019decoupled} for 20B tokens: a 50M-token linear warmup (100M at sizes M and L), a plateau at $6\times10^{-4}$, and linear decay over the final 2B tokens to $6\times10^{-5}$; betas $(0.9,0.95)$, weight decay 0.1, gradient clipping 1.0, bf16 autocast, and \texttt{torch.compile}. Global batch size is $96\times4096$ tokens on 8 H100s (one node or 2$\times$4) for width 768, M, and L, 2 GPUs for XS, and 1 GPU for XXS (8 and 16 accumulation steps); micro-batch size is 6, except 3 at L and 1 for the 32K continuation. Per-rank data order depends on GPU count and is therefore fixed within each size.

\textbf{Seeds.} Seed 1 is 1337 for every model. Seed 2 is 1338, except for the width-768 Transformer and SPS (seed 2); tied-head SPS uses seeds 2 and 3. Because the chunk-shuffle seed matches the model seed, seed-2 comparisons between those two models and the separated models do not share data order.

\begin{table}[!htbp]
\centering
\scriptsize
\caption{Run families differ from the base recipe only in the listed fields (otherwise untied heads, legacy chunk sampler, full-sweep validation). mb: per-GPU micro-batch size in 4096-token sequences (8$\times$6 denotes 8 GPUs with 6 sequences each per step). Kernel: training attention kernel (``as above'': Triton for the Transformer and SPS, flex otherwise). FW-Edu x60: extended corpus (App.~\ref{app:isoflop}). Seq.: Sequential.}
\label{tab:protocol}
\setlength{\tabcolsep}{2pt}
\begin{tabular*}{\linewidth}{@{\extracolsep{\fill}}>{\raggedright\arraybackslash}p{3.4cm}>{\raggedright\arraybackslash}p{2.3cm}lllll@{}}
\toprule
Family & Data / tokens & Warmup & \shortstack{GPUs\\$\times$ mb} & Seeds (s1; s2) & Sampler & Kernel \\
\midrule
S Transformer, SPS (untied) & FW-Edu / 20B & 50M & 8$\times$6 & 1337; 2 & legacy & Triton \\
S tied Transformer / SPS & FW-Edu / 20B & 50M & 8$\times$6 & 1337; SPS 2, 3 & legacy & Triton \\
S separated main and controls & FW-Edu / 20B & 50M & 8$\times$6 & 1337; 1338 & legacy & flex \\
Added controls (YOCO family, 24L, widened 12+6, shared Seq.\ untied, multi-level) & FW-Edu / 20B & 50M & 8$\times$6 & 1337 & legacy & flex; 24L Triton \\
Ladder XXS (3+3, width 384) & FW-Edu / 20B & 50M & 1$\times$6 & 1337; 1338 & legacy & as above \\
Ladder XS (4+4, 512) & FW-Edu / 20B & 50M & 2$\times$6 & 1337; 1338; 1339 & legacy & as above \\
Ladder M (12+12, 1024) & FW-Edu / 20B & 100M & 8$\times$6 & 1337; 1338 & legacy & as above \\
L Sequential 18+18 (1280) & FW-Edu / 20B & 100M & 8$\times$3 & 1337 & legacy & flex \\
Iso-FLOP S, M & FW-Edu x60 / 36.5B, 38.5B & 50M, 100M & 8$\times$6 & 1337 & balanced & flex \\
32K continuation (6 arms) & PG19 + FW-Edu long / 1.05B & 26M & 8$\times$1 & 1337 & balanced & as above \\
\bottomrule
\end{tabular*}
\end{table}

\textbf{Data-order notes.} In-training validation runs every 250M tokens for the Transformer and SPS and every 1B tokens for separated models (validation data only). The legacy chunked sampler repeats about 4.28\% of one rank's stream (0.5\% of all tokens), identically across all 20B runs. When a multi-GPU run resumes from a checkpoint, all ranks adopt rank 0's start offset, shifting ranks 1--7 by at most 805k tokens per chunk (under 0.1\%); single-GPU runs are unaffected.

\subsection{Evaluation Protocols}
\label{app:evalproto}
\paragraph{Validation loss.} Validation uses bf16 on non-overlapping 4096-token sequences: 13.1M scored tokens in the full sweep and 512 sequences (2,094,978 tokens) in the paired sub-sweeps used for interventions. No-op controls change loss by at most 0.0001 nats/token. An intervention magnitude counts as replicated when both seeds agree in sign and differ by at most 1.5$\times$. Every reported validation loss is the full sweep of the final checkpoint.

The 0.005 tie threshold corresponds to roughly 2.4 pooled seed standard deviations, estimated from eight two-seed arms and three tied-head SPS runs (10 degrees of freedom; chi-square 95\% CI $[0.0014, 0.0036]$ under common variance; the untied Transformer contributes about 36\% of the sum of squares). Tables round seed ranges, while differences in the text use unrounded losses. For \S\ref{sec:eff}, independent arm means give Sequential 6+6 minus SPS an approximate 95\% $t_{10}$ interval of $[0.0002, 0.0094]$ around $+0.0048$, and a gain-recovery delta-method interval of about $\pm6.5$ points (both ignoring same-seed covariance). Because the point estimate falls below the 0.005 threshold, we treat the two as tied even though the interval excludes zero. For single-seed runs we call a difference only beyond $\max(\text{seed spread}, 0.005)$.

\paragraph{Downstream tasks.} LAMBADA (OpenAI), HellaSwag, PIQA, ARC-Easy, ARC-Challenge, WinoGrande, OpenBookQA and SciQ are scored zero-shot with the lm-eval 0.4.9.1 task definitions, through our own loader and forward pass under bf16 autocast. The metric is length-normalized accuracy (divided by the answer's characters), except plain accuracy for LAMBADA and WinoGrande. Gold bits per byte is the NLL of lm-eval's gold continuation string (for LAMBADA the final word, for WinoGrande the remainder after the blank) per UTF-8 byte. GPT-2 (124M) through our path reproduces lm-eval's own backend on every task to the printed digit, and a per-model self-test verifies causality and reproduces the reported validation loss. Intervals are percentile bootstraps over examples (10,000 resamples per task; macro = mean of resampled per-task statistics) after averaging each example over seeds; they cover example sampling, not training noise.

\paragraph{Out-of-domain sets and retrieval.} Each out-of-domain set (C4, WikiText-103, Python code, PG19, arXiv abstracts; FineWeb-Edu validation as reference) is about 5M tokens in 4K windows, scored in bits per byte. Passkey hides a five-digit key at seven depths in filler of length 1K, 2K or 4K (100 trials per cell) and scores whether the argmax over number tokens equals the key. Multi-key NIAH hides several key--value needles and scores 5-way candidate accuracy (chance 0.2).

\paragraph{32K continued pretraining.} Each size-S main model (Transformer, SPS, Two-tower and Sequential at 6+6 and 12+12) continues from its seed-1 20B checkpoint for 1.05B tokens at context 32,768 with one recipe: PG19 training books and FineWeb-Edu documents of $\ge$32,769 tokens from unused shards (2.99B tokens; 0.35 epoch), document masking, RoPE base $10^4\to5\times10^5$, fresh AdamW at $3\times10^{-4}$ with 26M-token warmup and linear decay to $3\times10^{-5}$ over the final 10\%, batch $16\times32{,}768$ on 8 GPUs. SPS keeps its 64-token prediction window over 65,536 slots. A continuation costs 4.67$\times10^{18}$ FLOPs for the Transformer and 6+6 models and 9.1$\times10^{18}$ for SPS and 12+12 models, about 20\% of pretraining. Evaluation uses 79 PG19 test books and 100 held-out FineWeb-Edu documents (first 32,768 tokens each), plus passkey and multi-key NIAH from 2K to 32K (7 depths, 50 trials).

\section{Mechanistic Analyses}
\label{app:mech}
\label{app:roles}
\label{app:readaccess}
Methods, controls and full results behind the stream roles (\S\ref{sec:circuit}) and the read tests (\S\ref{sec:needs}), in that order, followed by their seed and size-M replication.

\subsection{Methods and Control Gates}
\label{app:methods}
\paragraph{Circuit scores.} Induction lift compares attention to matching keys with uniform attention over the causally visible keys, summing matches in the scored key stream and counting only queries with an allowed match. Stream means average all heads. Ablation effects are scored on induction tokens: targets $x_{t+1}$ with an earlier $x_j=x_{t+1}$ and $x_{j-1}=x_t$, where $j\le t-64$ excludes SPS's prediction-slot window.

\paragraph{Head-ablation controls.} Each pool draws three distinct non-target heads 100 times: random draws from any such heads, pattern-matched draws from the five nearest unused heads in standardized entropy, log attended distance, first-token mass and output norm, excluding high previous-token heads. Specificity is the loss of top-1 accuracy on induction tokens the intact model predicts correctly, with $p=(k+1)/101$ ($k$: control draws at least as costly as the target), Holm-corrected at $\alpha=0.05$ over the four tests per model and seed. The all-layer pools, metric and $p$-value were fixed before measurement; the layer-1 exclusions were added afterward. Neither pool matches the target's all-token cost, so the tests measure selectivity for induction tokens, not equal overall damage.

\paragraph{Patching and path tracing.} Patching uses duplicated 256-token segments and scores mean answer-versus-distractor logits over 96 examples. Path tracing decomposes matching-minus-control QK scores by source and requires residual reconstruction relative error $\le$0.01 and QK absolute error $\le$1.0.

\paragraph{Read restrictions, probes, lesions.} Read reach keeps the nearest 16, 64, 256 or 1024 state positions and masks or mean-replaces farther keys, with native-NLL and untrained-initialization controls; the unrestricted read reproduces native loss exactly. Linear next-, current- and previous-token probes train on positions 0--511 and evaluate on 512--1023 (every fourth position; standardized features; top 8192 tokens plus ``other''; Adam, 12 epochs), against trained-embedding, untrained-architecture, unigram and bigram baselines (bigram 5.241 / 5.236 nats/token for next / previous token); final next-token probes must do worse than the LM head. Lesions zero a layer's, attention's or MLP's residual contribution in one stream (SPS: the slot rows); mean ablation uses clean per-channel means from 256 held-out sequences, and the unlesioned control reproduces full-sweep loss.

\subsection{Circuit Statistics and Head-Ablation Controls}
\label{app:induction}
\label{app:controls}
Tab.~\ref{tab:circuit-full} extends Tab.~\ref{tab:circuit} to every model and seed. Across models and seeds, the roles act as a division of labor rather than a strict partition: in Sequential 9+3 and top-aligned Two-tower 12+6, state has more heads with lift above 3 than prediction; Two-tower 12+12's state has nearly as many; and AF-SPS reverses the previous-token ordering. On synthetic repeated random tokens, prediction still holds the largest lifts across all four separated models (maximum prediction lift 122--824 versus state 0.7--26.1 across seeds 1/2).

\begin{table}[t]
\centering
\scriptsize
\caption{In every Two-tower and Sequential model and seed, state queries carry more previous-token mass and prediction queries the higher mean lift. Lift $>3$: heads (of 12 per layer). S/P: state/prediction queries; SPS rows split key slots. $p$: uncorrected, random/matched controls. --: not run.}
\label{tab:circuit-full}
\setlength{\tabcolsep}{1.8pt}
\begin{tabular*}{\linewidth}{@{\extracolsep{\fill}}>{\raggedright\arraybackslash}p{3.9cm}ccccccc}
\toprule
Model & Prev.\ mass & Mean lift & Lift $>3$ & \multicolumn{2}{c}{$\Delta$NLL, abl./ctrl.} & \multicolumn{2}{c}{$p$, rand./match.} \\
\cmidrule(lr){5-6}\cmidrule(lr){7-8}
 & S/P & S/P & S/P & induct.\ tok. & all tok. & all layers & excl.\ L1 \\
\midrule
Transformer, tied head & 0.75/-- & 4.03/-- & 30/-- & 0.334/0.061 & 0.107/0.043 & 0.010/0.010 & 0.010/0.010 \\
\addlinespace
Transformer (s1) & 0.74/-- & 3.74/-- & 31/-- & 0.276/0.026 & 0.087/0.042 & 0.030/0.010 & 0.030/0.010 \\
Transformer (s2) & 0.75/-- & 3.61/-- & 34/-- & 0.324/0.010 & 0.105/0.009 & 0.010/0.010 & 0.010/0.010 \\
\addlinespace
SPS, tied head (s2), state keys & 0.31/0.18 & 2.03/4.62 & 11/37 & \multirow{2}{*}{0.099/0.000} & \multirow{2}{*}{0.030/0.003} & \multirow{2}{*}{0.079/0.198} & \multirow{2}{*}{0.030/0.198} \\
\quad pred.\ keys & 0.94/0.26 & 5.12/5.25 & 107/110 &  &  &  &  \\
SPS, tied head (s3), state keys & 0.67/0.15 & 2.45/5.56 & 17/40 & \multirow{2}{*}{0.297/0.001} & \multirow{2}{*}{0.096/0.002} & \multirow{2}{*}{0.010/0.010} & \multirow{2}{*}{0.010/0.010} \\
\quad pred.\ keys & 0.77/0.27 & 5.11/5.16 & 97/108 &  &  &  &  \\
\addlinespace
SPS (s1), state keys & 0.34/0.43 & 2.12/4.28 & 15/35 & \multirow{2}{*}{0.076/0.010} & \multirow{2}{*}{0.019/0.005} & \multirow{2}{*}{0.099/0.287} & \multirow{2}{*}{0.030/0.010} \\
\quad pred.\ keys & 0.64/0.36 & 3.91/4.44 & 88/98 &  &  &  &  \\
SPS (s2), state keys & 0.17/0.50 & 2.39/5.45 & 10/37 & \multirow{2}{*}{0.165/0.003} & \multirow{2}{*}{0.051/0.002} & \multirow{2}{*}{0.040/0.257} & \multirow{2}{*}{0.010/0.010} \\
\quad pred.\ keys & 0.95/0.31 & 4.20/4.54 & 92/95 &  &  &  &  \\
\addlinespace
AF-SPS & 0.25/0.49 & 1.11/1.12 & 0/0 & 0.022/0.002 & 0.063/0.005 & -- & -- \\
\addlinespace
Two-tower 12+12 (s1) & 0.91/0.38 & 2.96/4.38 & 30/34 & 0.072/0.029 & 0.026/0.017 & 0.050/0.010 & 0.050/0.010 \\
Two-tower 12+12 (s2) & 0.80/0.37 & 3.07/4.38 & 29/30 & 0.118/0.000 & 0.038/0.013 & 0.010/0.010 & 0.010/0.010 \\
\addlinespace
Two-tower 12+12 (shared attn. only) & 0.77/0.35 & 2.45/3.30 & 22/19 & 0.303/0.048 & 0.120/0.022 & -- & -- \\
\addlinespace
Two-tower 6+6 (s1) & 0.79/0.41 & 3.49/4.58 & 11/21 & 0.202/0.021 & 0.066/0.017 & 0.198/0.653 & 0.010/0.010 \\
Two-tower 6+6 (s2) & 1.00/0.37 & 3.53/4.76 & 17/20 & 0.101/0.019 & 0.049/0.012 & 0.248/0.792 & 0.069/0.010 \\
\addlinespace
Two-tower 12+6 (top-aligned) & 0.89/0.30 & 3.01/5.85 & 30/23 & 0.082/0.008 & 0.042/0.006 & -- & -- \\
\addlinespace
Sequential 12+12 (s1) & 0.71/0.20 & 2.06/4.69 & 12/40 & 0.150/0.009 & 0.055/0.007 & 0.040/0.010 & 0.020/0.010 \\
Sequential 12+12 (s2) & 0.71/0.14 & 2.17/4.77 & 11/44 & 0.124/0.012 & 0.037/0.010 & 0.010/0.010 & 0.010/0.010 \\
\addlinespace
Shared Sequential 12+12 (s1) & 0.91/0.17 & 2.04/4.97 & 10/45 & 0.245/0.016 & 0.092/0.016 & -- & -- \\
Shared Sequential 12+12 (s2) & 0.84/0.21 & 1.99/5.43 & 12/49 & 0.294/0.095 & 0.105/0.047 & -- & -- \\
\addlinespace
Sequential 3+9 & 0.66/0.18 & 2.22/4.95 & 4/33 & 0.342/0.025 & 0.142/0.036 & -- & -- \\
\addlinespace
Sequential 6+6 (s1) & 0.76/0.21 & 3.20/5.25 & 15/23 & 0.273/0.015 & 0.130/0.018 & 0.010/0.436 & 0.010/0.010 \\
Sequential 6+6 (s2) & 0.87/0.21 & 3.13/5.10 & 10/28 & 0.131/$-$0.001 & 0.086/0.018 & 0.079/0.386 & 0.079/0.010 \\
\addlinespace
Sequential 6+6 ($\langle$predict$\rangle$ input) & 0.75/0.27 & 2.90/5.45 & 12/26 & 0.257/0.029 & 0.141/0.015 & -- & -- \\
\addlinespace
Sequential 6+6 (tied head, $\langle$predict$\rangle$ input) (s1) & 0.80/0.32 & 3.58/4.74 & 11/22 & 0.231/0.032 & 0.112/0.020 & -- & -- \\
Sequential 6+6 (tied head, $\langle$predict$\rangle$ input) (s2) & 0.81/0.30 & 3.54/5.03 & 12/24 & 0.187/0.025 & 0.066/0.030 & -- & -- \\
\addlinespace
Sequential 9+3 & 0.81/0.16 & 3.27/5.98 & 24/13 & 0.251/0.003 & 0.077/0.012 & -- & -- \\
\addlinespace
Shared Two-tower 12+12 & 0.75/0.22 & 2.16/4.07 & 13/29 & 0.180/0.027 & 0.061/0.032 & 0.059/0.010 & 0.010/0.010 \\
\bottomrule
\end{tabular*}
\end{table}

\begin{figure}[htbp]
\centering
\includegraphics[width=0.8\linewidth]{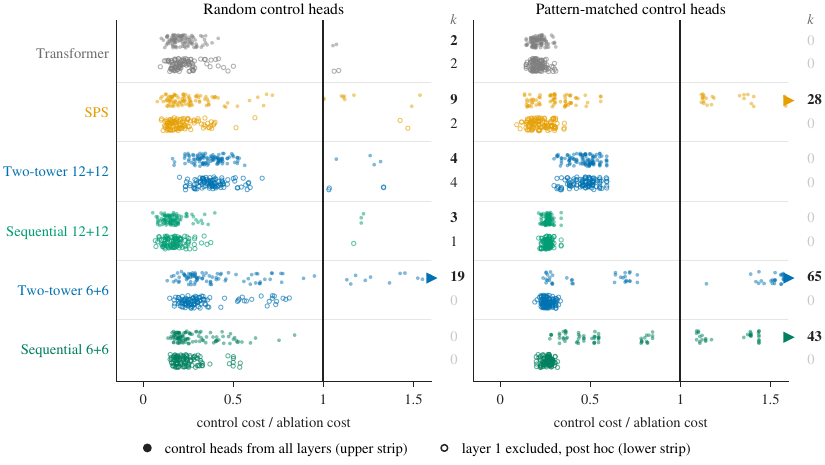}
\caption{In SPS and the 6+6 separated models, the costly controls are layer-1 heads. Seed-1 control/target cost in induction-token accuracy; line: 1; triangles: above 1.6; $k$: controls at least as costly as the target. Filled strips: all layers; open: layer 1 excluded.}
\label{fig:app-ctrl-dist}
\end{figure}

Fig.~\ref{fig:app-ctrl-dist} shows why layer 1 affects the ablation controls in \S\ref{sec:circuit}: in SPS and the 6+6 separated models the costliest controls are layer-1 heads, and excluding them drops Two-tower 6+6's pattern-matched $k$ from 65 to 0. Per-seed Holm results appear in Tab.~\ref{tab:circuit} (where Two-tower 6+6's low-mass control pool has only two distinct heads per seed).

\subsection{Activation Patching and Path Tracing}
\label{app:patching}
Patching tests whether a single state head carries enough signal for retrieval (Tab.~\ref{tab:patching}). The corruption swaps two unique tokens with different successors in the first copy of a duplicated segment, moving the token immediately before the match from the answer to a distractor while keeping the query and token inventory fixed. Restoring the strongest previous-token head at the answer slot (control: a donor from another example) recovers part of the logit difference in the separated models, whereas AF-SPS and the shared-weight models recover little. Because recovery does not rank the separated models consistently, patching supports the stream split rather than an ordering among models.

\begin{table}[t]
\centering
\footnotesize
\caption{One previous-token state head restores part of the logit difference in the separated models; random donors restore almost none. Seeds 1/2 (tied-head SPS: 2/3), or seed 1 with a 95\% bootstrap interval.}
\label{tab:patching}
\setlength{\tabcolsep}{4pt}
\begin{tabular*}{\linewidth}{@{\extracolsep{\fill}}lll}
\toprule
Model & Head patched in & Random donor \\
\midrule
Transformer & 0.50 / 0.26 & $-$0.04 / $-$0.01 \\
Transformer, tied head & 0.48 {\scriptsize[0.14, 0.90]} & $-$0.01 {\scriptsize[$-$0.04, 0.01]} \\
\midrule
AF-SPS & 0.00 {\scriptsize[$-$0.01, 0.01]} & 0.00 {\scriptsize[$-$0.03, 0.00]} \\
SPS & 0.01 / 0.16 & 0.00 / $-$0.01 \\
SPS, tied head & 0.10 / 0.02 & 0.00 / 0.01 \\
\midrule
Two-tower 6+6 & 0.23 / 0.09 & $-$0.02 / $-$0.10 \\
Two-tower 12+6 (top-aligned) & 0.10 {\scriptsize[0.03, 0.21]} & 0.02 {\scriptsize[$-$0.01, 0.04]} \\
Two-tower 12+12 (shared attn. only) & 0.04 {\scriptsize[0.00, 0.14]} & 0.02 {\scriptsize[0.00, 0.06]} \\
Two-tower 12+12 & 0.10 / 0.37 & 0.00 / $-$0.02 \\
\midrule
Sequential 3+9 & 0.29 {\scriptsize[0.15, 0.61]} & $-$0.02 {\scriptsize[$-$0.04, 0.03]} \\
Sequential 6+6 & 0.50 / 0.23 & 0.00 / $-$0.04 \\
Sequential 6+6 ($\langle$predict$\rangle$ input) & 0.73 {\scriptsize[0.55, 0.83]} & 0.00 {\scriptsize[$-$0.13, 0.04]} \\
Sequential 6+6 (tied head, $\langle$predict$\rangle$ input) & 0.19 / 0.08 & $-$0.02 / 0.00 \\
Sequential 9+3 & 0.38 {\scriptsize[0.10, 0.86]} & $-$0.01 {\scriptsize[$-$0.03, 0.00]} \\
Sequential 12+12 & 0.35 / 0.34 & $-$0.05 / $-$0.08 \\
\midrule
Shared Two-tower 12+12 & 0.00 {\scriptsize[$-$0.01, 0.01]} & $-$0.01 {\scriptsize[$-$0.01, 0.00]} \\
Shared Sequential 12+12 & 0.02 / 0.00 & 0.02 / 0.00 \\
\bottomrule
\end{tabular*}
\end{table}

Path tracing provides an observational counterpart. Two-tower 12+12 (shared attn.\ only), Shared Two-tower 12+12, AF-SPS and Shared Sequential 12+12 fail the exactness gate. In the remaining models, 3--12 state heads carry 80\% of the state-head QK score, the largest single contribution is 30--67\%, and attention sources account for most of the signal (85--91\% in Two-tower, except 67\% in Two-tower 6+6 seed 2, where an MLP is the largest source). Key-slot patching in Two-tower 12+12 (seed 1) gives the causal analogue: restoring all heads' key slots recovers 77\%, versus 25\% for the three previous-token heads alone; matching draws on many state heads.

\subsection{Role Probes and Baselines}
\label{app:probes}
Comparing the probes of \S\ref{sec:circuit} and \S\ref{sec:why} against untrained and bigram baselines (Tab.~\ref{tab:probe_floors}, seeds 1/2) separates what training adds from what the architecture preserves. Trained state retains far more previous-token information than untrained state, while current-token readability falls with depth even at initialization. Meanwhile, state predicts the next token near the bigram baseline (last-layer next-token probe 5.35 for SPS state and 4.81--5.06 for separated state, versus 3.69 for the Transformer, 3.5--3.7 for prediction streams, and 5.24 for bigram), so the retained information is specific to the previous token. Final prediction probes stay above the LM head's loss, so linear probes miss part of what the model uses.

\begin{table}[t]
\centering
\footnotesize
\caption{Trained state retains much more information about the previous token than untrained state. Last-layer linear-probe loss (nats/token); the Transformer uses its single stream for both. Untrained entries without $\pm$: one initialization.}
\label{tab:probe_floors}
\setlength{\tabcolsep}{4pt}
\begin{tabular*}{\linewidth}{@{\extracolsep{\fill}}lllll}
\toprule
 & \multicolumn{2}{c}{Next token (prediction)} & \multicolumn{2}{c}{Previous token (state)} \\
\cmidrule(lr){2-3}\cmidrule(lr){4-5}
Model & trained & untrained & trained & untrained \\
\midrule
Transformer & 3.69$_{\pm.01}$ & 7.16$_{\pm<.01}$ & 3.98$_{\pm.01}$ & 7.11$_{\pm<.01}$ \\
SPS & 3.61$_{\pm.01}$ & 7.57$_{\pm<.01}$ & 2.70$_{\pm.30}$ & 7.19$_{\pm<.01}$ \\
Two-tower 12+12 & 3.52$_{\pm<.01}$ & 7.56$_{\pm<.01}$ & 2.93$_{\pm.06}$ & 7.09$_{\pm<.01}$ \\
Sequential 12+12 & 3.51$_{\pm<.01}$ & 7.39 & 2.03$_{\pm<.01}$ & 7.09 \\
Two-tower 6+6 & 3.64$_{\pm.01}$ & 7.73 & 2.09$_{\pm.12}$ & 7.13 \\
Sequential 6+6 & 3.63$_{\pm.02}$ & 7.73 & 1.95$_{\pm.03}$ & 7.13 \\
\bottomrule
\end{tabular*}
\end{table}

\subsection{Read Reach}
\label{app:readreach}
Tab.~\ref{tab:readdepth_full} (top) reports the recent-position restriction across all two-stream models (main models seeds 1/2, tied-head SPS seeds 2/3, single-run controls seed 1) and the Transformer. In the four main separated models, mean-replacing distant state at $N{=}64$ costs 0.34--0.63 nats/token versus 0.32--0.40 for masking---at least as much in every model except Two-tower 6+6 (0.34 versus 0.40), which rules out a masking artifact. Restricting every layer of the Transformer at $N{=}64$ costs a similar amount (untied $0.364\pm0.030$, tied 0.342; mean replacement 0.617/0.517 and 0.656).

\begin{table}[t]
\centering
\scriptsize
\caption{Every two-stream model needs distant state. Loss increase (nats/token) from restricting prediction to recent positions (top) or to the first/last $m$ state levels (middle/bottom; 12-level towers: rows $\leq k$ keep $m=k$, rows $\geq k$ keep $m=13-k$). $^{\ast}$: native read not same-depth. --: unchanged read or beyond state depth. No $\pm$: one seed.}
\label{tab:readdepth_full}
\setlength{\tabcolsep}{2pt}
\sbox0{%
\begin{tabular}{lrrrrrrrr}
\toprule
 & \multicolumn{4}{c}{masked, $N{=}$} & \multicolumn{4}{c}{mean-replaced, $N{=}$} \\
\cmidrule(lr){2-5}\cmidrule(lr){6-9}
Model & 16 & 64 & 256 & 1024 & 16 & 64 & 256 & 1024 \\
\midrule
AF-SPS & 1.37 & 0.41 & 0.083 & 0.006 & 1.33 & 0.80 & 0.26 & 0.013 \\
SPS & 0.98$_{\pm.47}$ & 0.53$_{\pm.30}$ & 0.18$_{\pm.12}$ & 0.034$_{\pm.028}$ & 1.63$_{\pm.04}$ & 0.91$_{\pm.23}$ & 0.36$_{\pm.22}$ & 0.073$_{\pm.065}$ \\
SPS, tied head & 1.32$_{\pm.35}$ & 0.82$_{\pm.27}$ & 0.33$_{\pm.19}$ & 0.078$_{\pm.059}$ & 1.35$_{\pm.16}$ & 1.02$_{\pm.17}$ & 0.58$_{\pm.17}$ & 0.15$_{\pm.10}$ \\
\midrule
Two-tower 6+6 & 1.04$_{\pm.01}$ & 0.40$_{\pm.02}$ & 0.091$_{\pm.016}$ & 0.009$_{\pm.003}$ & 0.71$_{\pm.01}$ & 0.34$_{\pm.01}$ & 0.11$_{\pm<.01}$ & 0.010$_{\pm.002}$ \\
Two-tower 12+6 (top-aligned) & 1.07 & 0.37 & 0.069 & 0.006 & 0.79 & 0.40 & 0.11 & 0.009 \\
Two-tower 12+12 (shared attn. only) & 1.02 & 0.36 & 0.069 & 0.006 & 0.86 & 0.42 & 0.11 & 0.009 \\
Two-tower 12+12 & 0.97$_{\pm<.01}$ & 0.36$_{\pm<.01}$ & 0.068$_{\pm<.001}$ & 0.006$_{\pm<.001}$ & 0.78$_{\pm<.01}$ & 0.40$_{\pm<.01}$ & 0.12$_{\pm<.01}$ & 0.008$_{\pm<.001}$ \\
\midrule
Sequential 3+9 & 1.28 & 0.36 & 0.070 & 0.006 & 1.10 & 0.51 & 0.12 & 0.009 \\
Sequential 6+6 & 1.24$_{\pm.02}$ & 0.37$_{\pm.01}$ & 0.070$_{\pm.001}$ & 0.007$_{\pm<.001}$ & 0.96$_{\pm.02}$ & 0.46$_{\pm.01}$ & 0.11$_{\pm<.01}$ & 0.010$_{\pm<.001}$ \\
Sequential 6+6 ($\langle$predict$\rangle$ input) & 0.84 & 0.30 & 0.067 & 0.006 & 0.78 & 0.34 & 0.093 & 0.008 \\
Sequential 6+6 (tied head, $\langle$predict$\rangle$ input) & 0.96$_{\pm.03}$ & 0.33$_{\pm<.01}$ & 0.071$_{\pm.001}$ & 0.007$_{\pm<.001}$ & 0.81$_{\pm.02}$ & 0.37$_{\pm.01}$ & 0.10$_{\pm.01}$ & 0.009$_{\pm<.001}$ \\
Sequential 9+3 & 1.21 & 0.50 & 0.10 & 0.007 & 0.71 & 0.37 & 0.12 & 0.011 \\
Sequential 12+12 & 1.17$_{\pm.13}$ & 0.32$_{\pm.03}$ & 0.068$_{\pm.002}$ & 0.007$_{\pm<.001}$ & 1.29$_{\pm.15}$ & 0.63$_{\pm.13}$ & 0.14$_{\pm.01}$ & 0.011$_{\pm<.001}$ \\
\midrule
Shared Two-tower 12+12 & 1.07 & 0.33 & 0.063 & 0.006 & 1.14 & 0.55 & 0.12 & 0.009 \\
Shared Sequential 12+12 & 0.95$_{\pm.01}$ & 0.32$_{\pm.01}$ & 0.067$_{\pm.002}$ & 0.006$_{\pm<.001}$ & 0.95$_{\pm.01}$ & 0.43$_{\pm.01}$ & 0.098$_{\pm.004}$ & 0.007$_{\pm<.001}$ \\
\bottomrule
\end{tabular}}%
\ifdim\wd0>\linewidth\resizebox{\linewidth}{!}{\usebox0}\else
\begin{tabular*}{\linewidth}{@{\extracolsep{\fill}}lrrrrrrrr}
\toprule
 & \multicolumn{4}{c}{masked, $N{=}$} & \multicolumn{4}{c}{mean-replaced, $N{=}$} \\
\cmidrule(lr){2-5}\cmidrule(lr){6-9}
Model & 16 & 64 & 256 & 1024 & 16 & 64 & 256 & 1024 \\
\midrule
AF-SPS & 1.37 & 0.41 & 0.083 & 0.006 & 1.33 & 0.80 & 0.26 & 0.013 \\
SPS & 0.98$_{\pm.47}$ & 0.53$_{\pm.30}$ & 0.18$_{\pm.12}$ & 0.034$_{\pm.028}$ & 1.63$_{\pm.04}$ & 0.91$_{\pm.23}$ & 0.36$_{\pm.22}$ & 0.073$_{\pm.065}$ \\
SPS, tied head & 1.32$_{\pm.35}$ & 0.82$_{\pm.27}$ & 0.33$_{\pm.19}$ & 0.078$_{\pm.059}$ & 1.35$_{\pm.16}$ & 1.02$_{\pm.17}$ & 0.58$_{\pm.17}$ & 0.15$_{\pm.10}$ \\
\midrule
Two-tower 6+6 & 1.04$_{\pm.01}$ & 0.40$_{\pm.02}$ & 0.091$_{\pm.016}$ & 0.009$_{\pm.003}$ & 0.71$_{\pm.01}$ & 0.34$_{\pm.01}$ & 0.11$_{\pm<.01}$ & 0.010$_{\pm.002}$ \\
Two-tower 12+6 (top-aligned) & 1.07 & 0.37 & 0.069 & 0.006 & 0.79 & 0.40 & 0.11 & 0.009 \\
Two-tower 12+12 (shared attn. only) & 1.02 & 0.36 & 0.069 & 0.006 & 0.86 & 0.42 & 0.11 & 0.009 \\
Two-tower 12+12 & 0.97$_{\pm<.01}$ & 0.36$_{\pm<.01}$ & 0.068$_{\pm<.001}$ & 0.006$_{\pm<.001}$ & 0.78$_{\pm<.01}$ & 0.40$_{\pm<.01}$ & 0.12$_{\pm<.01}$ & 0.008$_{\pm<.001}$ \\
\midrule
Sequential 3+9 & 1.28 & 0.36 & 0.070 & 0.006 & 1.10 & 0.51 & 0.12 & 0.009 \\
Sequential 6+6 & 1.24$_{\pm.02}$ & 0.37$_{\pm.01}$ & 0.070$_{\pm.001}$ & 0.007$_{\pm<.001}$ & 0.96$_{\pm.02}$ & 0.46$_{\pm.01}$ & 0.11$_{\pm<.01}$ & 0.010$_{\pm<.001}$ \\
Sequential 6+6 ($\langle$predict$\rangle$ input) & 0.84 & 0.30 & 0.067 & 0.006 & 0.78 & 0.34 & 0.093 & 0.008 \\
Sequential 6+6 (tied head, $\langle$predict$\rangle$ input) & 0.96$_{\pm.03}$ & 0.33$_{\pm<.01}$ & 0.071$_{\pm.001}$ & 0.007$_{\pm<.001}$ & 0.81$_{\pm.02}$ & 0.37$_{\pm.01}$ & 0.10$_{\pm.01}$ & 0.009$_{\pm<.001}$ \\
Sequential 9+3 & 1.21 & 0.50 & 0.10 & 0.007 & 0.71 & 0.37 & 0.12 & 0.011 \\
Sequential 12+12 & 1.17$_{\pm.13}$ & 0.32$_{\pm.03}$ & 0.068$_{\pm.002}$ & 0.007$_{\pm<.001}$ & 1.29$_{\pm.15}$ & 0.63$_{\pm.13}$ & 0.14$_{\pm.01}$ & 0.011$_{\pm<.001}$ \\
\midrule
Shared Two-tower 12+12 & 1.07 & 0.33 & 0.063 & 0.006 & 1.14 & 0.55 & 0.12 & 0.009 \\
Shared Sequential 12+12 & 0.95$_{\pm.01}$ & 0.32$_{\pm.01}$ & 0.067$_{\pm.002}$ & 0.006$_{\pm<.001}$ & 0.95$_{\pm.01}$ & 0.43$_{\pm.01}$ & 0.098$_{\pm.004}$ & 0.007$_{\pm<.001}$ \\
\bottomrule
\end{tabular*}\fi
\par\medskip
\begin{tabular*}{\linewidth}{@{\extracolsep{\fill}}lrrrrrr}
\toprule
 & \multicolumn{6}{c}{levels $\leq k$, $k{=}$} \\
\cmidrule(lr){2-7}
Model & 2 & 4 & 6 & 8 & 10 & 12 \\
\midrule
Two-tower 6+6 & 1.21$_{\pm.01}$ & 0.39$_{\pm.04}$ & -- & -- & -- & -- \\
Two-tower 12+6 (top-aligned)$^{\ast}$ & 1.82 & 1.54 & 1.39 & 1.39 & 1.39 & 1.39 \\
Two-tower 12+12 (shared attn. only) & 2.04 & 1.48 & 1.06 & 0.54 & 0.16 & -- \\
Two-tower 12+12 & 2.17$_{\pm.05}$ & 1.60$_{\pm.05}$ & 0.97$_{\pm.03}$ & 0.36$_{\pm.01}$ & 0.16$_{\pm<.01}$ & -- \\
\midrule
Sequential 3+9$^{\ast}$ & 0.62 & -- & -- & -- & -- & -- \\
Sequential 6+6$^{\ast}$ & 1.47$_{\pm<.01}$ & 1.07$_{\pm.05}$ & 0.94$_{\pm.06}$ & -- & -- & -- \\
Sequential 6+6 ($\langle$predict$\rangle$ input)$^{\ast}$ & 1.29 & 0.98 & 0.88 & -- & -- & -- \\
Sequential 6+6 (tied head, $\langle$predict$\rangle$ input)$^{\ast}$ & 1.51$_{\pm.12}$ & 1.03$_{\pm.04}$ & 0.90$_{\pm.04}$ & -- & -- & -- \\
Sequential 9+3$^{\ast}$ & 2.25 & 2.16 & 2.16 & 2.16 & -- & -- \\
Sequential 12+12$^{\ast}$ & 4.07$_{\pm.83}$ & 2.82$_{\pm.50}$ & 2.66$_{\pm.52}$ & 2.58$_{\pm.53}$ & 2.56$_{\pm.57}$ & 2.44$_{\pm.50}$ \\
\bottomrule
\end{tabular*}
\par\medskip
\begin{tabular*}{\linewidth}{@{\extracolsep{\fill}}lrrrrrr}
\toprule
 & \multicolumn{6}{c}{levels $\geq k$, $k{=}$} \\
\cmidrule(lr){2-7}
Model & 2 & 4 & 6 & 8 & 10 & 12 \\
\midrule
Two-tower 6+6 & 0.010$_{\pm<.001}$ & 0.10$_{\pm.01}$ & 0.68$_{\pm.06}$ & -- & -- & -- \\
Two-tower 12+6 (top-aligned)$^{\ast}$ & 1.34 & 1.08 & 0.66 & 0.43 & 0.20 & 0.31 \\
Two-tower 12+12 (shared attn. only) & 0.059 & 0.063 & 0.11 & 0.14 & 0.28 & 0.75 \\
Two-tower 12+12 & 0.034$_{\pm.022}$ & 0.063$_{\pm.033}$ & 0.094$_{\pm.038}$ & 0.18$_{\pm.03}$ & 0.32$_{\pm.01}$ & 0.66$_{\pm.04}$ \\
\midrule
Sequential 6+6$^{\ast}$ & 0.86$_{\pm.06}$ & 0.44$_{\pm.05}$ & -- & -- & -- & -- \\
Sequential 6+6 ($\langle$predict$\rangle$ input)$^{\ast}$ & 0.79 & 0.38 & -- & -- & -- & -- \\
Sequential 6+6 (tied head, $\langle$predict$\rangle$ input)$^{\ast}$ & 0.78$_{\pm.05}$ & 0.35$_{\pm.02}$ & -- & -- & -- & -- \\
Sequential 9+3$^{\ast}$ & 2.03 & 1.33 & 0.72 & 0.31 & -- & -- \\
Sequential 12+12$^{\ast}$ & 2.67$_{\pm.48}$ & 2.82$_{\pm.74}$ & 2.67$_{\pm.73}$ & 2.38$_{\pm.77}$ & 1.98$_{\pm1.07}$ & -- \\
\bottomrule
\end{tabular*}
\end{table}

\subsection{Read Depth under Evaluation-Time Restriction}
\label{app:readdepth}
For \S\ref{sec:read}, prediction is restricted at evaluation to the first or last $m$ state levels while keeping trained K/V projections and all positions (read maps $\min(\ell,m)$ and $\max(\ell,a-m+1)$; in Two-tower 12+12 the sweep pairs the last $m$ levels with the first $m{+}1$). Because the model was trained to read every level, these tests measure what the trained model relies on rather than which single level a model trained from scratch would prefer; keeping only early levels also changes what later prediction layers read. Several bounds move layers off their trained levels (the starred models and the extremes: limiting Two-tower 6+6 to levels $\geq6$ yields a final-state read, and Sequential 12+12 to $\leq12$ a same-depth read costing 2.44), which is why \S\ref{sec:final} retrains models from scratch. SPS, AF-SPS, and Shared Two-tower 12+12 read state-layer inputs, and Shared Sequential 12+12's shared projections were never trained on substituted levels, so they are omitted. To test level preference under training rather than evaluation-time substitution, App.~\ref{app:mixweights} lets prediction learn its own mixture over levels.

\subsection{What Prediction Reads When It Can Choose}
\label{app:mixweights}
As a trained counterpart to evaluation-time restrictions, we let every prediction layer of Sequential 12+12 read a learned mixture $r_\ell=\sum_{l=1}^{12} w_{\ell l}\,s^l$ of all state levels, initialized one-hot on the final level (144 extra parameters, $+0.03\%$ FLOPs). Tab.~\ref{tab:mixweights} reports each level group's norm share after 20B tokens ($\|w_{\ell l}s^l\|$ summed over the group divided by the sum over all levels, averaged over 1M validation tokens and over prediction layers in each third). The final level remains the largest single source in every layer (0.35--0.64 of the read), matching Sequential's choice, while later prediction layers draw roughly a third of their norm share from levels 1--9 and pay the highest cost when forced back to the final level alone. In terms of block outputs ($s^l-s^{l-1}$), 56--58\% of mixture mass sits on the first three blocks (58\% at S, 56\% at M). The evaluation-time finding that later levels matter more (\S\ref{sec:read}) therefore does not imply that earlier levels are unused; App.~\ref{app:multilevel} reports the resulting loss gains across sizes.

\begin{table}[!htbp]
\centering
\footnotesize
\caption{The final level is the largest source of every learned read. Sequential 12+12, seed 1: norm share of the read (unsigned magnitude of each level's contribution) by level group and prediction-layer third; $\Delta$NLL (nats/token, range over layers) when one layer's mixture is replaced by the final level.}
\label{tab:mixweights}
\setlength{\tabcolsep}{4pt}
\begin{tabular*}{\linewidth}{@{\extracolsep{\fill}}llrrrrr@{}}
\toprule
 & Prediction layers & Level 12 & Levels 10--11 & Levels 4--9 & Levels 1--3 & $\Delta$NLL, final only \\
\midrule
S & 1--4 & 0.56 & 0.21 & 0.14 & 0.10 & 0.001--0.022 \\
 & 5--8 & 0.46 & 0.20 & 0.21 & 0.13 & 0.004--0.016 \\
 & 9--12 & 0.43 & 0.24 & 0.21 & 0.12 & 0.007--0.037 \\
\midrule
M & 1--4 & 0.53 & 0.18 & 0.16 & 0.12 & 0.002--0.016 \\
 & 5--8 & 0.51 & 0.18 & 0.19 & 0.12 & 0.001--0.011 \\
 & 9--12 & 0.39 & 0.26 & 0.23 & 0.12 & 0.013--0.033 \\
\bottomrule
\end{tabular*}
\end{table}

\subsection{Layer and Module Lesions}
\label{app:lesions}
To verify that the layer profiles in Fig.~\ref{fig:read-depth}b--d (seeds 1/2; SPS mean ablation seed 1) do not depend on the ablation method, Tab.~\ref{tab:lesion_mean} compares zero ablation with mean ablation. Both methods rank layers similarly but differ in absolute magnitude, since mean replacement removes position-specific structure while injecting a nonzero vector; relative layer importance is therefore more robust than raw cost. Sequential 12+12's final state layer remains costly under mean ablation (2.38/0.71 in seeds 1/2), well above Two-tower 12+12's (0.08/0.09). At 6+6, no state layer after layer 1 is cheap (minimum 0.148 in Two-tower and 0.236 in Sequential, versus 0.062 and 0.047 at 12+12), with both attention and MLP contributing (Fig.~\ref{fig:app-depth-split}), so cheap interior layers are a property of depth.

\begin{table}[t]
\centering
\footnotesize
\caption{Zero and mean ablation rank layers alike. Spearman $\rho$ between the two layer profiles, and loss increase (nats/token) from lesioning layer 1 under each. SPS: one seed.}
\label{tab:lesion_mean}
\setlength{\tabcolsep}{4pt}
\begin{tabular*}{\linewidth}{@{\extracolsep{\fill}}lrrrrrr}
\toprule
 & \multicolumn{2}{c}{Spearman $\rho$} & \multicolumn{2}{c}{Layer 1, state} & \multicolumn{2}{c}{Layer 1, prediction} \\
\cmidrule(lr){2-3}\cmidrule(lr){4-5}\cmidrule(lr){6-7}
Model & state & pred. & zero & mean & zero & mean \\
\midrule
SPS & 0.97 & 0.98 & 3.45 & 0.80 & 4.28 & 3.27 \\
Two-tower 12+12 & 0.97$_{\pm<.01}$ & 0.97$_{\pm.02}$ & 4.59$_{\pm.32}$ & 8.33$_{\pm.78}$ & 0.45$_{\pm.01}$ & 0.42$_{\pm<.01}$ \\
Sequential 12+12 & 0.89$_{\pm.01}$ & 0.93$_{\pm.07}$ & 3.85$_{\pm.18}$ & 6.16$_{\pm2.68}$ & 0.76$_{\pm.01}$ & 0.57$_{\pm.03}$ \\
Two-tower 6+6 & 0.97$_{\pm.03}$ & 0.86$_{\pm.09}$ & 5.71$_{\pm.34}$ & 6.60$_{\pm.55}$ & 0.86$_{\pm.05}$ & 0.50$_{\pm.01}$ \\
Sequential 6+6 & 1.00$_{\pm<.01}$ & 0.91$_{\pm.03}$ & 3.79$_{\pm.08}$ & 4.03$_{\pm.02}$ & 2.05$_{\pm.20}$ & 0.86$_{\pm.08}$ \\
\bottomrule
\end{tabular*}
\end{table}

\subsection{Seed Replication}
\label{app:seeds}
Under the 1.5$\times$ rule (App.~\ref{app:evalproto}), effects that \textbf{replicate in sign and magnitude} are $N{=}64$ reach in the separated models and the Transformer, layer-1 lesions, first-$m$ restrictions in Two-tower 12+12 and Sequential 6+6, role probes, and Sequential 12+12's head ablations. \textbf{Same sign, varying magnitude:} SPS head ablation and reach, Sequential 12+12's top-state-layer lesion (2.26/0.56), and several patching effects. \textbf{Not replicated:} head-ablation specificity outside Sequential 12+12. The main-text claims on reach, read depth and Sequential 12+12's head specificity rest on replicating effects; lesion and patching magnitudes support direction only.

\subsection{Replication at Size M}
\label{app:interp-m}
Tab.~\ref{tab:interp-m} repeats the mechanistic analyses of \S\S\ref{sec:circuit}--\ref{sec:arch} (excluding patching and path tracing) on both seeds of 327M Sequential (12+12, width 1024). All qualitative patterns replicate at size M: state retains previous-token attention and probe readability to its top layer, prediction holds the higher induction lift and relies on distant state, and the three-head ablation passes all Holm-corrected pools without layer-1 exclusion. As at width 768, interior state layers are cheap to lesion at 12+12 depth, while level-restriction ratios and final-layer lesion costs vary across seeds (and because width changes with size, these numbers do not isolate scale alone).

\begin{table}[!htbp]
\centering
\scriptsize
\caption{Mechanistic findings replicate on both seeds of 327M (size M) Sequential 12+12. Values for size M are seed 1 / seed 2; width-768 comparators are Sequential 12+12 (and 6+6 where noted). Three-head ablation at M is 1.6\% of state heads and passes all four control pools ($p\le0.02$, Holm-significant without layer-1 exclusion; low-mass control selectivity 0.54/0.95$\times$). At 6+6's head fraction (8 heads, 4.2\%), seed 2's pattern-matched test requires layer-1 exclusion, as in 6+6. Untrained state prev.-token probe at M: 7.24; bigram baseline: 5.24.}
\label{tab:interp-m}
\setlength{\tabcolsep}{3pt}
\begin{tabular*}{\linewidth}{@{\extracolsep{\fill}}lll@{}}
\toprule
Analysis / metric & M Sequential 12+12 (s1 / s2) & Width-768 comparator \\
\midrule
Peak prev.-token mass (S vs.\ P) & 0.83/0.92 vs.\ 0.19/0.21 & 0.71 vs.\ 0.17 (12+12) \\
Mean induction lift (P vs.\ S) & 4.57/4.49 vs.\ 2.04/2.70 & 4.73 vs.\ 2.12 (12+12) \\
3-head ablation: $\Delta$NLL / selectivity & 0.081/0.081 / 3.83/2.45$\times$ & 2.74/3.39$\times$ (12+12) \\
8-head ablation selectivity (4.2\% heads) & 2.06/1.64$\times$ & 1.53--2.10$\times$ (6+6, 3 heads) \\
$N{=}64$ read-reach cost ($\Delta$NLL) & 0.331 / 0.305 & 0.29--0.38 (12+12, 6+6) \\
State top-layer probe: prev.\ / next NLL & 1.99/1.98 / 5.13/4.94 & 2.03 / 4.81--5.06 (sep.) \\
Min.\ interior state lesion ($\Delta$NLL) & 0.031 / 0.041 & $\approx$0.05~(12+12), \mbox{$\approx$0.24~(6+6)} \\
First-4 vs.\ last-3 level cost ratio & 2.6 / 7.4$\times$ & 1.1 / 2.5$\times$ (12+12) \\
Final state-layer lesion ($\Delta$NLL) & 0.47 / 0.11 & 2.26 / 0.56 (12+12) \\
Gain share vs.\ Two-tower: first-occ.\ / ind. & 65.5\% / 6.7\% & 58--61\% / 8--14\% \\
\bottomrule
\end{tabular*}
\end{table}

\section{Architecture Controls}
\label{app:archctrl}
\label{app:variants}
Controls for \S\ref{sec:arch} (prediction capacity and depth, tower ratio, a 24-layer baseline, YOCO's design choices, prediction input, and weight sharing), followed by the token breakdown for \S\ref{sec:why}. Parameters and FLOPs of every arm are in Tab.~\ref{tab:roster_full}.

\subsection{Prediction Depth versus Capacity}
\label{app:depthcap}
\label{app:archabl}
\textbf{MLP reallocation.} Shifting MLP width from state to prediction leaves Two-tower 12+12 at 2.737, versus 2.733 originally and 2.702 for Sequential 12+12 (seed 1): extra prediction capacity at equal cost does not substitute for the final-state read.

\textbf{Top-aligned Two-tower 12+6} (six prediction layers reading state levels 7--12) reaches 2.757, behind Two-tower 12+12 (2.733), but it also uses 26\% fewer non-embedding parameters and 23\% fewer FLOPs. Since widening cannot match both budgets simultaneously, we widened the prediction MLP twice (seed 1 each). Matched to Two-tower 12+12's parameters (MLP 5824; 184.1M, 10\% fewer FLOPs), it reaches 2.722, 0.010 below Two-tower 12+12. Matched to its FLOPs (MLP 8576; 21\% more parameters), it reaches 2.699, 0.034 below Two-tower 12+12 and level with Sequential 12+12 (2.702). At matched parameters, a six-layer prediction tower reading later state levels beats a twelve-layer one reading same-level state with fewer FLOPs (one seed), though Sequential 12+12 still leads the parameter-matched 12+6 by 0.021.

\subsection{Tower Ratio}
\label{app:ratio}
\begin{figure}[htbp]
\centering
\includegraphics[width=0.85\linewidth]{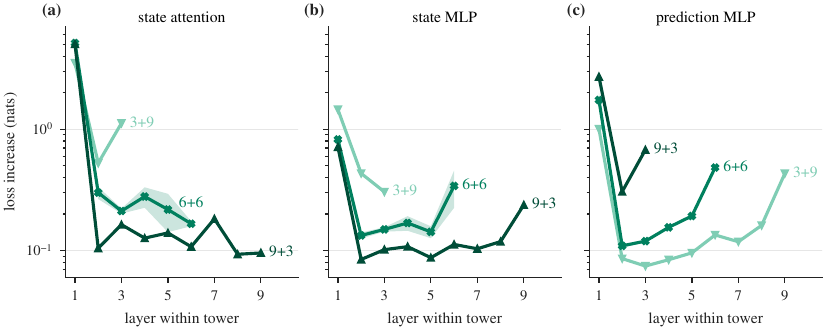}
\caption{Every layer of the shortened towers contributes. Module lesion costs (nats/token, log scale) by layer: (a) state attention, (b) state MLP, (c) prediction MLP. Bands: two-seed range (6+6).}
\label{fig:app-depth-split}
\end{figure}

At fixed total depth and width, an even split gives the lowest loss (Tab.~\ref{tab:ratio}): Sequential 6+6 (2.801) beats 3+9 (2.813) and 9+3 (2.815), and every layer of the shortened towers contributes (Fig.~\ref{fig:app-depth-split}). Additional 1:2 and 2:1 splits show the same pattern at both sizes, with state-heavy splits beating state-light ones (at S, 8+4 ties 6+6). Towers here reach at most 9 layers, and unequal splits inherit initialization scaled to the longer tower (output projections $0.02/\sqrt{2\max(a,b)}$), which we did not vary. The even split also lets every prediction layer read the final state, as Two-tower's last layer alone does.

\begin{table}[!htbp]
\centering
\footnotesize
\caption{The even split has the lowest loss. Sequential's final validation loss by state:prediction split at fixed $a+b$, width, parameters and FLOPs. One seed per split; $^{\ast}$: two-seed mean.}
\label{tab:ratio}
\begin{tabular}{@{}lccccc@{}}
\toprule
State:prediction & 1:3 & 1:2 & 1:1 & 2:1 & 3:1 \\
\midrule
XXS ($a+b=6$) & -- & 3.251 & 3.232$^{\ast}$ & 3.239 & -- \\
S ($a+b=12$) & 2.813 & 2.808 & 2.801$^{\ast}$ & 2.804 & 2.815 \\
\bottomrule
\end{tabular}

\end{table}

\subsection{Transformer 24L at the 12+12 Budget}
\label{app:tf24}
A 24-layer width-768 Transformer matches the 12+12 models in non-embedding parameters (184.1M) and FLOPs (0.751 GFLOPs/token) and reaches 2.756 (seed 1), behind Two-tower 12+12 (2.733) and Sequential 12+12 (2.702). The 12+12 models therefore do not win merely from having twice the 12-layer Transformer's parameters: a separate prediction stream helps even with a same-depth read, and the final-state read adds another 0.031. Macro zero-shot accuracy follows the same order: \dsTrTwentyFourMacro\% for Transformer 24L versus \dsTTTwelveMacro\% for Two-tower 12+12 and \dsSeqTwelveMacro\% for Sequential 12+12 (App.~\ref{app:downstream}).

\subsection{Relation to YOCO}
\label{app:yoco}
Sequential is closest to YOCO's cross-decoder \citep{sun2024yoco}, which also reads the final state at every upper layer, but differs in two respects: YOCO continues the lower residual stream into the upper layers (whereas Sequential starts prediction fresh) and computes a single shared K/V set from the final state (whereas Sequential gives each prediction layer its own K/V projection). Tab.~\ref{tab:yoco} varies each choice at 6+6 while keeping full causal attention in the lower stack to isolate the cross-decoder.

\begin{table}[!htbp]
\centering
\scriptsize
\caption{Both of YOCO's choices cost loss at equal FLOPs. All 6+6 with full attention below; $\Delta$: relative to Sequential 6+6. New rows: one seed.}
\label{tab:yoco}
\setlength{\tabcolsep}{2pt}
\begin{tabular*}{\linewidth}{@{\extracolsep{\fill}}lllrrrrr@{}}
\toprule
Model & Prediction stream & K/V of read & \shortstack{non-emb.\\(M)} & \shortstack{total\\(M)} & GFLOPs & NLL & $\Delta$ \\
\midrule
Sequential 6+6 (2 seeds) & fresh, own embedding & per layer & 92.0 & 207.9 & 0.414 & 2.801 & -- \\
Sequential 6+6, $\langle$predict$\rangle$ input & fresh, one token & per layer & 92.0 & 169.3 & 0.414 & 2.805 & $+0.004$ \\
Shared K/V, width-restored & fresh, own embedding & one set & 92.3 & 208.2 & 0.415 & 2.816 & $+0.015$ \\
Residual continuation & state's final residual & per layer & 92.0 & 169.3 & 0.414 & 2.826 & $+0.025$ \\
Residual continuation, detached & same, no gradient & per layer & 92.0 & 169.3 & 0.414 & 2.846 & $+0.044$ \\
YOCO-style cross-decoder & state's final residual & one set & 86.1 & 163.4 & 0.402 & 2.850 & $+0.049$ \\
Transformer (2 seeds) & -- & -- & 92.0 & 169.3 & 0.414 & 2.865 & $+0.064$ \\
\bottomrule
\end{tabular*}
\end{table}

Both YOCO choices hurt loss at equal FLOPs. Continuing the state residual costs 0.025 against Sequential 6+6 and 0.021 against $\langle$predict$\rangle$-input Sequential (which matches its 169.3M total parameters), ruling out the embedding table as the cause. Sharing one K/V set costs 0.015 even after restoring parameters (prediction MLP widened to 2752), while the full YOCO-style model---which has 6.4\% fewer non-embedding parameters by design---finishes 0.049 behind Sequential 6+6 and 0.014 ahead of the Transformer. The input feeding the fresh stream barely matters (2.801 versus 2.805); what matters is running prediction as a separate stream with per-layer K/V projections. These controls also clarify why Two-tower 6+6 beats the Transformer by 0.027 at equal cost: every 6+6 model here dedicates six layers to cross-attending over state, and every one beats the Transformer by 0.014 (YOCO-style, at lower cost) to 0.064, including residual continuation (+0.039 at equal cost). Two-tower 6+6 sits within that band, and at the larger budget Two-tower 12+12 beats the 24-layer Transformer by 0.023 (App.~\ref{app:tf24}). At these sizes, dedicating half the layers to prediction with direct state reads outweighs the lost single-stream depth (one seed per new row).

\subsection{Prediction Input}
\label{app:predinput}
Sequential's prediction embedding table gives it more total parameters than the Transformer and SPS, but replacing it with a single learned $\langle$predict$\rangle$ token stays within the tie threshold (seed 1: 2.805 versus 2.803). This variant matches the Transformer's 169.3M total parameters and Sequential 6+6's macro accuracy (\dsSeqSixPredMacro\% versus \dsSeqSixMacro\%). Swapping the input only at inference costs 0.935--1.333 nats/token, so each model relies on whichever input it was trained with. At size M, a $\langle$predict$\rangle$-input Sequential 12+12 with an untied head matches the M Transformer's 430.2M total parameters and reaches 2.609 (seed 1), versus 2.611 for M Sequential with its own table (two seeds) and 2.653 for the M Transformer (head tying: App.~\ref{app:tiedgroup}).

\subsection{Weight Sharing}
\label{app:sharing}
Under weight sharing (as in SPS), the tied pair does not settle the read preference because it also differs in K/V sharing, which App.~\ref{app:yoco} shows is costly on its own. Shared Two-tower 12+12 (seed 1, reads $\ell{-}1$) and Shared Sequential 12+12 (seeds 1/2, reads final state through an extra K/V projection of 1.18M parameters and 2.4M FLOPs/token) run SPS's stack twice without interleaving or a prediction window, with a tied head and $\langle$predict$\rangle$ input. An untied Shared Sequential 12+12 uses its own prediction table and reads the final state through each shared layer's own K/V weights (seed 1; 92.0M non-embedding parameters, 0.751 GFLOPs/token). The tied pair favors Shared Two-tower by 0.012 (2.814 versus 2.826; already 2.970 versus 2.988/2.987 before decay at 22.4k PFLOPs), whereas untied Shared Sequential reaches 2.806---0.009 below Shared Two-tower and 0.021 below tied Shared Sequential. Because tied Shared Sequential reads the final state through a single K/V set (a 0.015 cost in App.~\ref{app:yoco}, comparable to its 0.012 deficit), the ordering follows head, input, and K/V choices rather than the read alone; an untied Shared Two-tower matched in all three would be needed to isolate the read under weight sharing. Sharing attention alone also hurts Two-tower 12+12 (2.733 to 2.767). Weight sharing erases most of the benefit of separation: the best shared model trails SPS by 0.009 and merely ties Sequential 6+6 at 1.8$\times$ its FLOPs. Weight sharing also weakens causal retrieval localization: Shared Two-tower 12+12's head ablation depends on the control pool and single-head patching recovers nothing (Tabs.~\ref{tab:circuit-full}, \ref{tab:patching}), so the circuit split in \S\ref{sec:circuit} requires separate weights.

\subsection{Where Sequential's Gain Falls}
\label{app:finalread}
Breaking down Sequential's gain over Two-tower (\S\ref{sec:why}) by repeat distance (Tab.~\ref{tab:finalread}; separated models, seeds 1/2) shows that every bucket improves across all four seed pairings at both depths with similar relative gains. First-occurrence tokens account for 59--63\% of the gain at 12+12 and 55--61\% at 6+6 in every pairing, while induction tokens (12,651 qualifying tokens requiring a preceding-bigram match) take less than their token share in every seed, with gain share $(n_{\rm ind}/n_{\rm all})(\Delta_{\rm ind}/\Delta_{\rm all})$ of 9.9\%/14.4\% at 12+12 and 8.2\%/10.6\% at 6+6. (The repeat-distance breakdown uses the 512-sequence paired sub-sweep of 2.09M tokens, so its overall $\Delta$ differs slightly from Tab.~\ref{tab:2x2}'s full sweep; the induction sweep uses 16 sequences, 65,520 tokens.) Against the Transformer, both SPS and Sequential 12+12 also improve on first-occurrence tokens---slightly less in relative terms than on repeated tokens (1.9--2.2\% versus 2.3--3.5\% for SPS; 5.1--5.4\% versus 5.7--6.8\% for Sequential)---with first-occurrence tokens carrying 53--55\% of SPS's gain and 58\% of Sequential's.

\begin{table}[!ht]
\centering
\footnotesize
\caption{Sequential's gain over Two-tower is not concentrated on copyable tokens. By how far back the target last occurred: token share, Sequential minus Two-tower loss $\Delta$ (nats/token), relative difference and share of gain. Bottom: induction tokens' share of gain.}
\label{tab:finalread}
\label{tab:token-gain}
\setlength{\tabcolsep}{4pt}
\begin{tabular*}{\linewidth}{@{\extracolsep{\fill}}lrrrrrrr}
\toprule
 & & \multicolumn{3}{c}{12+12} & \multicolumn{3}{c}{6+6} \\
\cmidrule(lr){3-5}\cmidrule(lr){6-8}
Last occurrence & tokens & $\Delta$ & rel. & gain & $\Delta$ & rel. & gain \\
\midrule
First-occurrence & 42.8\% & $-0.043$ & $-1.1$\% & 61\% & $-0.050$ & $-1.2$\% & 58\% \\
1--16 back & 13.9\% & $-0.012$ & $-0.9$\% & 5\% & $-0.017$ & $-1.2$\% & 6\% \\
17--64 & 21.0\% & $-0.017$ & $-1.2$\% & 12\% & $-0.022$ & $-1.5$\% & 13\% \\
65--256 & 14.5\% & $-0.026$ & $-1.3$\% & 12\% & $-0.032$ & $-1.5$\% & 13\% \\
257--1024 & 6.6\% & $-0.035$ & $-1.3$\% & 8\% & $-0.042$ & $-1.4$\% & 8\% \\
$>$1024 & 1.2\% & $-0.040$ & $-1.2$\% & 2\% & $-0.063$ & $-1.7$\% & 2\% \\
\midrule
all & 100\% & $-0.030$ & & & $-0.036$ & & \\
\midrule
Induction tokens & 19.3\% & \multicolumn{6}{c}{share of gain 8--14\% (both depths, all seed pairs)} \\
\bottomrule
\end{tabular*}
\end{table}

\textbf{Prediction depth.} Sequential's prediction stream becomes next-token readable much earlier than Two-tower's because it reads fully processed state from layer 1: Two-tower minus Sequential probe loss at 12+12 is 0.245--0.280 at prediction layer 1, 0.887--1.018 at layer 4, 0.136--0.148 at layer 11, and 0.005--0.009 at layer 12 (6+6: 0.763--0.811 at layer 1, $-0.006$ to 0.041 at layer 6; ranges over seed pairings). Probe loss reaches 5.0 and 4.5 nats/token at prediction layers 2 and 5 in Sequential 12+12 versus 6 and 8 in Two-tower 12+12, and at layers 2 and 3 in Sequential 6+6 versus 3 and 4 in Two-tower 6+6 (Transformer: 5.0 at layer 7).

\section{Efficiency, Equal Compute and Scaling}
\label{app:scaling}
Completes \S\ref{sec:eff}: Sequential versus SPS at equal training FLOPs, in wall-clock time, with matched heads and against AF-SPS; then the size ladder, why the margins narrow, and the multi-level read that recovers part of the narrowing.

\subsection{Equal Compute versus SPS}
\label{app:isoflop}
\label{app:compute}
At 20B tokens Sequential and SPS have equal non-embedding parameters, but SPS spends 1.83$\times$ (S) and 1.92$\times$ (M) the training FLOPs, so we compare them per training FLOP.

\paragraph{Measured iso-FLOP runs.} We trained Sequential 6+6 and M Sequential 12+12 to exactly SPS's training FLOPs (36.50B and 38.45B tokens; relative difference under $3\times10^{-6}$) at the unchanged configuration, the compute-matched comparison \citet{monea2026sps} made for their Transformer. Sequential reaches 2.760 at S (SPS 2.797) and 2.557 at M (SPS 2.583), 0.037 and 0.025 lower (one seed each). These runs need more than the 27B-token training file, so they read an extended file (the original file as a byte-identical prefix plus 24 further shards of the same release, same validation set), use a balanced sampler (each rank owns five whole chunks), and decay over their last 10\% (3.65B and 3.85B tokens). On the shared constant-learning-rate plateau, at equal tokens, they sit 0.002 (S, 16B and 18B) and 0.004 (M, 18B) below the same-seed 20B runs and within 0.002 of the two-seed means, against margins of 0.025--0.037. They see 1.83$\times$ and 1.92$\times$ SPS's unique tokens, inherent to equal compute on a cheaper model. Per token (20B), SPS is level with Sequential at S (0.0048) and ahead at M by 0.028. This iso-FLOP comparison was pre-registered; \S\ref{sec:eff}'s head-matched comparison was run post hoc, after untied SPS was trained.

\paragraph{Interpolated multipliers.} Interpolating SPS's loss curve in log compute gives the FLOPs SPS needs to reach Sequential's final loss: 1.71, 1.79, 1.79 and 1.67$\times$ from XXS to M (M 5--95\% range under seed noise 1.59--1.75). These move along model size at fixed data, so the measured runs above are the direct test.

\paragraph{Compute matching.} Compute in Fig.~\ref{fig:frontier} is tokens $\times 3 \times$ forward FLOPs/token. The pre-decay comparison uses the 18B-token checkpoints of the Transformer and 6+6 models and interpolates SPS and the 12+12 models near 9.9B tokens in cumulative compute before averaging seeds; endpoints include each run's own decay. At the endpoints, seed-1/seed-2 Sequential 6+6 minus SPS gaps are $+0.0050/+0.0045$ (the seed-2 SPS run has a different data order), with 92\%/94\% gain recovery and 88--97\% across all seed pairings.

\subsection{Kernels and Same-Kernel Wall-Clock}
\label{app:wallclock}
Our Triton kernel skips only tiles above the causal diagonal and applies the prediction window, slot parity and document mask as dense biases, so for SPS it computes about half of a $(2T)^2$ score matrix per head while the FLOP count credits $2T(T{+}64)$ scores. To measure SPS with masked tiles skipped, we also time it on FlexAttention with block-sparse masks, storing slots as $[T$ state $\mid T$ prediction$]$ (a permutation of the same computation) so that the windowed prediction keys form a diagonal band and masked tiles are skipped. On the trained checkpoint the flex kernel matches the Triton loss within $3\times10^{-4}$ nats/token (gradient cosine $\ge0.9993$), and unit tests check the mask element by element.

\begin{table}[!htbp]
\centering
\footnotesize
\caption{On one kernel, SPS's step time exceeds Sequential 6+6's by less than its FLOP ratio. Training step time on one H100 (micro-batch $12\times4096$; median of 30 steps, three repetitions); MFU: model FLOPs ($3\times$ forward) over 989 TFLOP/s; K/V cache in bf16. The flex Transformer ran on another node.}
\label{tab:efficiency}
\setlength{\tabcolsep}{3pt}
\begin{tabular*}{\linewidth}{@{\extracolsep{\fill}}llrrrrrr@{}}
\toprule
Model & Kernel & ms/step & \shortstack{$\times$ Transf.\\(time)} & \shortstack{$\times$ Transf.\\(FLOPs)} & MFU & \shortstack{Peak mem.\\(GiB)} & \shortstack{K/V\\KiB/tok.} \\
\midrule
Transformer & Triton & 180.5 & 1.00 & 1.00 & 0.34 & 27.4 & 36 \\
SPS & Triton & 403.2 & 2.23 & 1.83 & 0.28 & 43.5 & 36 \\
\midrule
Transformer (other node) & flex & 143.3 & 1.00 & 1.00 & 0.43 & -- & 36 \\
SPS, interleaved, tile 128 & flex & 342.2 & 2.39 & 1.83 & 0.33 & -- & 36 \\
SPS, split, tile 128 & flex & 269.1 & 1.88 & 1.83 & 0.42 & -- & 36 \\
SPS, split, tile 64 & flex & 235.4 & 1.64 & 1.83 & 0.48 & 48.4 & 36 \\
Sequential 6+6 & flex & 134.5 & 0.94 & 1.00 & 0.46 & 30.7 & 36 \\
Two-tower 12+12 & flex & 240.1 & 1.68 & 1.81 & 0.47 & 50.6 & 72 \\
\bottomrule
\end{tabular*}
\end{table}

Each step is timed with \texttt{torch.compile}, forward, backward and fused AdamW, after 10 warmup steps; node-to-node variation is about $\pm4\%$ (Sequential 6+6: 134.5 versus 140.3 ms on two nodes). Decode speed is unmeasured, since timing it needs an incremental Two-tower decoder. On the same node, SPS (split, tile 64) takes 1.75$\times$ Sequential 6+6's step time, below its 1.83$\times$ FLOP ratio, and Two-tower 12+12 1.79$\times$ (Tab.~\ref{tab:efficiency}). Sequential 6+6 takes 0.94$\times$ the flex Transformer's time, within cross-node uncertainty, so we claim no speed advantage. Its 12 K/V sets (six state layers, one final-state projection per prediction layer) match the Transformer's cache. In the 32K continuation SPS on Triton took 3.2$\times$ the Transformer's step time, reflecting the kernel rather than its FLOPs (1.95$\times$ at 32K).

\subsection{Tied Heads}
\label{app:tiedgroup}
Head tying does not explain the efficiency result in \S\ref{sec:eff}. Tying helps the Transformer (2.862 to 2.851, seed 1) but hurts SPS (2.797 to 2.807, seeds 2/3). Tied-head Sequential 6+6 with $\langle$predict$\rangle$ input reaches 2.815/2.819---0.010 above tied-head SPS---and its per-seed gains over the tied-head Transformer (0.036/0.032) recover 82\%/72\% of tied-head SPS's 0.044, showing that the efficiency advantage persists with matched heads at slightly lower recovery.

\subsection{Almost-Free SPS}
\label{app:afsps}
AF-SPS \citep{langford2026free}, reimplemented as a cheaper efficiency reference (tied head, seed 1), shares the MLP between streams. Prediction uses a learned $\langle$predict$\rangle$ token without prediction-to-prediction attention; attention and norms are shared; for normalized residuals $\bar s,\bar p$ the gate is $g=\sigma(W_\text{in}[\bar s;\bar p])$ and the shared output $f=\text{MLP}(g\bar s+(1-g)\bar p)$, with per-stream output gates. Prediction layer $\ell$ reads state level $\ell{-}1$, as in SPS, and because state depends on prediction through the shared MLP a final-state read is impossible without changing the architecture. AF-SPS trades some of SPS's loss for lower compute (Tab.~\ref{tab:roster_full}).\footnote{Its 0.595 GFLOPs/token count one shared-MLP evaluation per position.} Our version differs from the original in scale (131M/20B versus 1B/100B), multi-head versus grouped-query attention, AdamW, dense attention, and context (4096 versus 8192), and omits tail-phasing and the two-pass kernel. Unlike the separated models, it has no head with induction lift above 3 and no induction-selective ablation (Tab.~\ref{tab:circuit-full}).

\subsection{Size Ladder}
\label{app:ladder}
We train the four main models at XXS, XS, S, and M (width 384, 512, 768, 1024; 6, 8, 12, and 24 Transformer layers; $a=b=L/2$, matching Transformer, Two-tower, and Sequential in non-embedding parameters and FLOPs/token), each for 20B tokens under the recipe of App.~\ref{app:training} with the same peak learning rate at every size (100M warmup at M as in \citealp{monea2026sps}; size S reuses the untied 6+6 runs). Forward GFLOPs/token from XXS to M are 0.100, 0.174, 0.414, and 1.165 for the Transformer, Two-tower, and Sequential, and 0.162, 0.298, 0.756, and 2.239 for SPS. At L (width 1280, 18+18, 767M non-embedding and 960M total parameters) we trained only Sequential (2.503, seed 1) and its multi-level variant; \citet{monea2026sps} report 2.537 for a Transformer and 2.470 for SPS of this shape under their own protocol, cited here only as reference points.

\begin{figure}[htbp]
\centering
\includegraphics[width=0.85\linewidth]{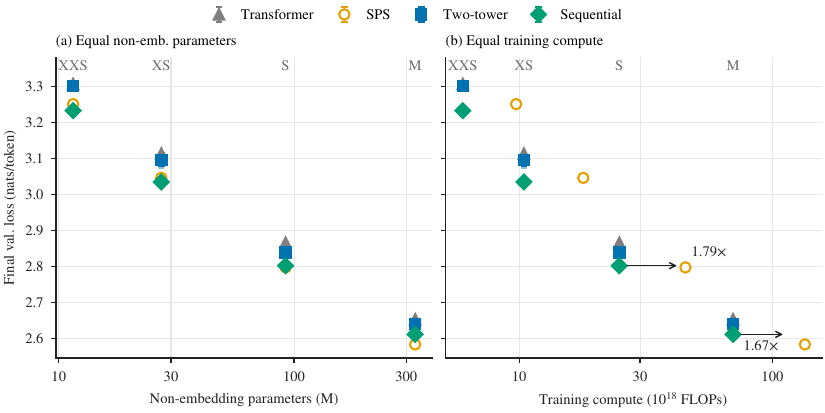}
\caption{Sequential leads the FLOP-matched models at every size. Loss against (a) non-embedding parameters and (b) training compute; arrows: FLOPs SPS needs to match Sequential. Bars: min--max over seeds (SPS open); dots: seeds.}
\label{fig:scaling}
\end{figure}

Sequential beats the Transformer and Two-tower at every size on the ladder (Tab.~\ref{tab:scaling}, Fig.~\ref{fig:scaling}). Two-tower needs 1.15--1.23$\times$ Sequential's FLOPs to match it, and the Transformer needs 1.24, 1.32, and 1.36$\times$ at XXS--S. At M the Transformer's multiplier must be extrapolated: 1.31$\times$ under the shared-$E$ fit, 1.28$\times$ under its own fit, and 1.23$\times$ extending the S--M segment linearly in log compute---consistent with either a flat or slowly declining multiplier (distinguishing them requires a Transformer above M). Sequential's macro accuracy also exceeds the Transformer's at every size, significantly at XS, S, and M (App.~\ref{app:downstream}).

\begin{table}[t]
\centering
\footnotesize
\caption{Sequential beats the FLOP-matched Transformer and Two-tower at every size. Final validation loss (20B tokens; $^\ddagger$: three seeds; no $\pm$: one). $N$: non-embedding parameters. FLOPs to match Seq.: compute a model needs to reach Sequential's final loss, relative to Sequential's ($^\dagger$: extrapolated). Seq.\ at SPS FLOPs: iso-FLOP run (App.~\ref{app:isoflop}). $^{\S}$: Sequential only.}
\label{tab:scaling}
\setlength{\tabcolsep}{0.6pt}
\begin{tabular*}{\linewidth}{@{\extracolsep{\fill}}lrccccrrrrc@{}}
\toprule
 & & \multicolumn{4}{c}{Final validation loss (nats/token)} & \multicolumn{1}{c}{Seq.$-$SPS} & \multicolumn{3}{c}{FLOPs to match Seq.\ ($\times$)} & \shortstack{Seq.\ at} \\
\cmidrule(lr){3-6}\cmidrule(lr){8-10}
Size & $N$ (M) & Transformer & SPS & Two-tower & Sequential & (nats) & Transf. & Two-tow. & SPS & \shortstack{SPS FLOPs} \\
\midrule
XXS & 11.5 & 3.306$_{\pm.003}$ & 3.250$_{\pm.001}$ & 3.301$_{\pm.004}$ & 3.232$_{\pm.002}$ & $-0.018$ & 1.24 & 1.20 & 1.71 & -- \\
XS & 27.3 & 3.113$_{\pm.001}$ & 3.045$_{\pm.001}$ & 3.095$_{\pm.016}^{\ddagger}$ & 3.034$_{\pm.002}^{\ddagger}$ & $-0.011$ & 1.32 & 1.23 & 1.79 & -- \\
S & 92.0 & 2.865$_{\pm.003}$ & 2.797$_{\pm.001}$ & 2.838$_{\pm<.001}$ & 2.801$_{\pm.002}$ & $+0.005$ & 1.36 & 1.21 & 1.79 & 2.760 \\
M & 327.2 & 2.653 & 2.583$_{\pm.002}$ & 2.639 & 2.611$_{\pm.005}$ & $+0.028$ & 1.31$^\dagger$ & 1.15$^\dagger$ & 1.67 & 2.557 \\
\midrule
L$^{\S}$ & 767 & -- & -- & -- & 2.503 & -- & -- & -- & -- & -- \\
\bottomrule
\end{tabular*}
\end{table}

\begin{figure}[htbp]
\centering
\includegraphics[width=0.62\linewidth]{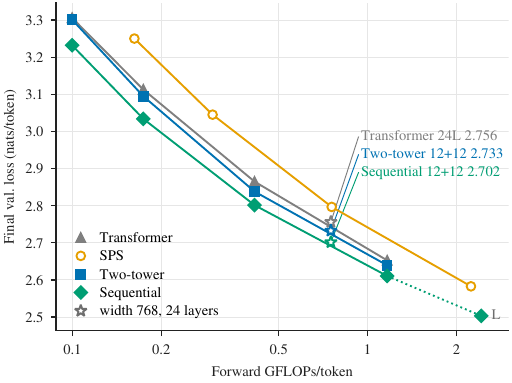}
\caption{At every forward-FLOP budget on the ladder, Sequential has the lowest loss. Final validation loss against forward GFLOPs/token (log scale), XXS to M. Stars: width-768 24-layer models; dotted: Sequential at L (one seed).}
\label{fig:fwdfrontier}
\end{figure}

\paragraph{Loss per forward FLOP.} Fig.~\ref{fig:fwdfrontier} plots final loss against forward FLOPs/token, where Sequential achieves the lowest loss at every budget on the ladder. Interpolated in log FLOPs, SPS's ladder reaches 2.958 at the size-S budget of 0.414 GFLOPs (Sequential 2.801, Transformer 2.865) and 2.711 at the size-M budget of 1.165 GFLOPs (Sequential 2.611, Transformer 2.653). Conversely, Sequential interpolated to SPS's FLOPs/token sits 0.191, 0.156, 0.106, and 0.068 below SPS from XXS to M (at M interpolated toward the one-seed L run). Along this axis Sequential has more non-embedding parameters than SPS at equal FLOPs, and interpolating along the ladder varies width and depth together. At fixed width 768 and 0.751 GFLOPs the measured order is Sequential 12+12 (2.702), Two-tower 12+12 (2.733), Transformer 24L (2.756), with S SPS at 2.797 and 0.756 GFLOPs.

\FloatBarrier
\subsection{Why the Margins Narrow}
\label{app:narrow}
From XXS to M, Sequential's lead over the Transformer moves from 0.074 and 0.079 to 0.064 and 0.042 nats/token; per token, SPS shifts from 0.018 behind Sequential at XXS to 0.028 ahead at M; and Sequential's lead over Two-tower falls from 0.069 at 3+3 to 0.029 at 12+12. Two measured factors account for roughly two-thirds of the narrowing between 92M and 327M (0.064 to 0.042, or 0.0215 before rounding): the depth split removes 0.009 and reading a single state level removes about 0.005 (note that the 24-layer Transformer and S multi-level control are single-seed, and the two M multi-level seeds differ by 0.019). Below we cover both factors, the fixed-token budget, and ruled-out explanations.

\begin{figure}[htbp]
\centering
\includegraphics[width=0.75\linewidth]{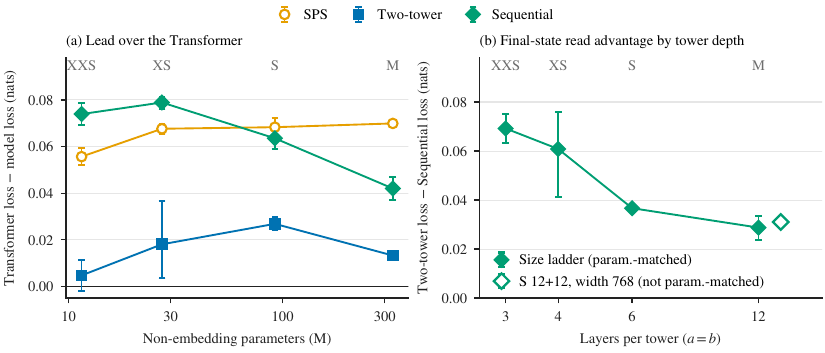}
\caption{SPS keeps its lead over the Transformer while the FLOP-matched separated models lose ground. (a) Transformer minus model final loss (nats/token) against non-embedding parameters. (b) Two-tower minus Sequential loss against layers per tower: ladder (filled), width-768 12+12 (open); bars: all seed pairings.}
\label{fig:depthgap}
\end{figure}

\paragraph{Depth split.} At equal FLOPs Sequential divides depth between two towers, whereas the Transformer, and SPS through weight sharing, apply full depth to every token. Because larger ladder sizes have more layers, splitting depth costs more there. At fixed width 768, the 12+12 models lead the 24-layer Transformer by 0.055 (Sequential) and 0.023 (Two-tower), compared to 0.064 and 0.027 for 6+6 over the 12-layer Transformer (App.~\ref{app:tf24}): doubling depth removes 0.009 of Sequential's lead, two-fifths of the S-to-M shrinkage (0.0215). SPS, which does not split depth, holds a steady 0.056--0.070 lead over the Transformer at every size while both depth-split models lose ground (Fig.~\ref{fig:depthgap}a). Layer by layer, M Sequential behaves like width-768 12+12 rather than 6+6 (App.~\ref{app:interp-m}).

\paragraph{Single read source.} Every Sequential prediction layer reads the same final state. Replacing that read with a learned mixture of levels (App.~\ref{app:multilevel}) helps more as towers deepen and recovers roughly a quarter of the narrowing: its lead over the Transformer falls from 0.071 at S to 0.055 at M, versus 0.064 to 0.042 for the final read. At L, multi-level Sequential reaches 2.489 (reference points: App.~\ref{app:ladder}). Sequential's margin over Two-tower narrows for a related reason (Fig.~\ref{fig:depthgap}b): deeper state towers enrich the intermediate levels Two-tower reads. At M, the next-token probe on prediction reaches 4.5 nats/token at 0.50 of the depth in Two-tower (one seed) and 0.46--0.50 in SPS, down from 0.57--0.61 at width 768, whereas Sequential remains at 0.34--0.35.

\paragraph{Token budget and compute.} All models train on 20B tokens (from roughly 1{,}700 down to 61 tokens per non-embedding parameter), moving along size at fixed data. Training length does not explain the smaller M margin: Sequential's lead over the Transformer builds more slowly at S and M (at 2B tokens, 55\% of its final size at S and 36--38\% at M), levels off by about 8B tokens at S and 12B at M, and then stays flat at M (0.037/0.048 for seeds 1/2 at 12B versus 0.037/0.047 at 20B). SPS's per-token lead at M costs 1.92$\times$ the training FLOPs, and at equal FLOPs Sequential beats SPS at both S and M (App.~\ref{app:isoflop}).

\paragraph{Ruled-out explanations.} Neither diminished role separation nor training instability explains the narrowing. The stream split replicates at M (App.~\ref{app:interp-m}), interior layers become cheap at 24 layers in every model including SPS, and SPS keeps its full margin. Training behavior is also consistent across models: the final decay lowers loss by the same 0.130--0.134 nats/token in all four M models, loss spikes recover within four steps, and every block's weight RMS settles at the same AdamW equilibrium across M models and their S comparators.

\paragraph{Caveats.} Learning rate and schedule were not tuned per size or model; width and depth change together (width per Transformer layer is 64, 64, 64, 43); and M two-seed standard deviations are 0.0022 (SPS) and 0.0069 (Sequential). The two extrapolations marked $^\dagger$ in Tab.~\ref{tab:scaling} use a shared-$E$ power-law fit $L=E+AN^{-\alpha}$ to all 16 cell means ($E=1.717$; every M mean within 0.005). Because $E$ is not well identified (1.47--2.16 across parameter and compute fits to XXS--S or to all sizes), we do not extrapolate beyond M.

\subsection{Multi-Level Read across Sizes}
\label{app:multilevel}
To test the single-read-source hypothesis, we replace the final-state read in the seed-1 Sequential configuration with a learned mixture $r_\ell=\sum_{l} w_{\ell l}\,s^l$, initialized one-hot on the final level (App.~\ref{app:mixweights}; at most 324 extra parameters). This read lowers loss by 0.005--0.014 at every size (Tab.~\ref{tab:multilevel}), with larger gains in deeper towers: 0.005--0.008 for 3- to 6-layer towers versus 0.012--0.014 for 12- to 18-layer towers, and 0.008 rising to 0.012 from 6+6 to 12+12 at width 768. The gain clears the single-seed threshold of App.~\ref{app:evalproto} at every size except M (and at XS by under 0.001); the two M seeds (2.607 and 2.588) differ by 0.019, so additional seeds are needed there. An input-dependent mixture reaches 2.694 at S 12+12, tying the static mixture. Reading only the final state thus becomes more restrictive as towers deepen, explaining part of the margin narrowing above; all main-text comparisons retain the plain final-state read.

\begin{table}[!htbp]
\centering
\footnotesize
\caption{The learned multi-level read lowers loss at every size. Final read: three seeds at XS, one at L; multi-level: seed 1 (M: two seeds).}
\label{tab:multilevel}
\setlength{\tabcolsep}{4pt}
\begin{tabular*}{\linewidth}{@{\extracolsep{\fill}}llrllr@{}}
\toprule
Size & Towers & $N$ (M) & Final read & Multi-level & $\Delta$ \\
\midrule
XXS & 3+3 & 11.5 & 3.232$_{\pm.002}$ & 3.224 & $-0.008$ \\
XS & 4+4 & 27.3 & 3.034$_{\pm.002}$ & 3.028 & $-0.005$ \\
S & 6+6 & 92.0 & 2.801$_{\pm.002}$ & 2.793 & $-0.008$ \\
S (width 768) & 12+12 & 184.1 & 2.702$_{\pm<.001}$ & 2.690 & $-0.012$ \\
M & 12+12 & 327.2 & 2.611$_{\pm.005}$ & 2.597$_{\pm.010}$ & $-0.013$ \\
L & 18+18 & 767 & 2.503 & 2.489 & $-0.014$ \\
\bottomrule
\end{tabular*}
\end{table}

\section{Beyond Validation Loss}
\label{app:beyond}
Full results behind \S\ref{sec:beyond}; protocols in App.~\ref{app:evalproto}.

\subsection{Downstream Tasks}
\label{app:downstream}
Tab.~\ref{tab:ds-acc} gives per-task accuracy and gold bits per byte, and Tab.~\ref{tab:ds-paired} gives paired bootstrap differences in macro scores against the same-size Transformer as well as direct comparisons against Two-tower and SPS (the third XS seed was not evaluated). Sequential beats Two-tower and the Transformer on bits per byte at every size from XXS to S, with intervals excluding zero in both seeds, and on macro accuracy wherever the gap is large enough to resolve (Sequential 12+12 versus Two-tower 12+12, Sequential 6+6 versus the Transformer, and XS and M versus both). In the seed average it never trails either model significantly (at M, bits per byte against the Transformer ties because seed 1 favors Sequential and seed 2 favors the Transformer, both significantly). Against SPS, Sequential trails significantly at S on bits per byte (accuracy tied) and at M on both ($-0.96$ accuracy points); SPS also leads at XXS. Per task, Sequential 6+6 trails Two-tower 6+6 on OBQA accuracy ($-2.2$ points) and ARC-E bits per byte, and M Sequential trails the M Transformer on ARC-E and SciQ bits per byte.

\begin{figure}[htbp]
\centering
\includegraphics[width=\linewidth]{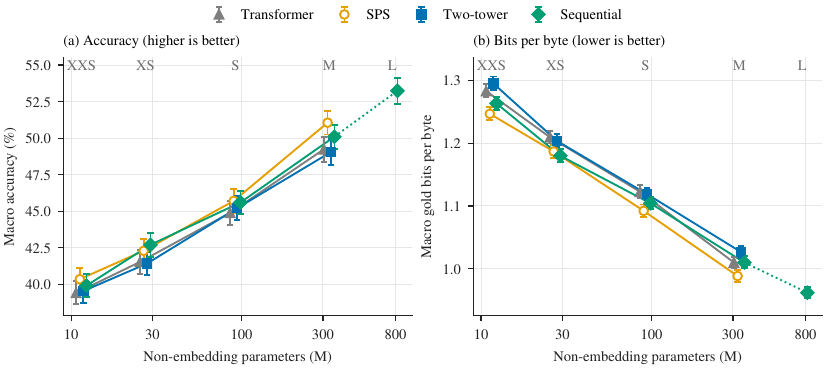}
\caption{Sequential's macro accuracy exceeds the Transformer's at every size. Macro accuracy (a) and gold bits per byte (b) against non-embedding parameters (one seed: M Transformer, M Two-tower, L). Bars: 95\% example-bootstrap intervals; dotted: Sequential at L.}
\label{fig:ds-scaling}
\end{figure}

Fig.~\ref{fig:ds-scaling} plots the macro scores of the four ladder models against non-embedding parameters. Sequential 18+18 at L reaches 53.2\% macro accuracy and 0.962 bits per byte (one seed).

\begin{table}[!htbp]
\centering
\caption{Sequential's macro gains over the Transformer come from several tasks, including LAMBADA and HellaSwag at every size. Accuracy (\%) / gold bits per byte per task, and macro scores. S: seeds averaged (also for Tab.~\ref{tab:ood}); $\langle$p$\rangle$: $\langle$predict$\rangle$ input.}
\label{tab:ds-acc}
\label{tab:ds-bpb}
{\scriptsize\setlength{\tabcolsep}{1pt}
\begin{tabular*}{\linewidth}{@{\extracolsep{\fill}}lccccccccccc@{}}
\toprule
Model & S & LAMB. & HS & PIQA & ARC-E & ARC-C & WG & OBQA & SciQ & \multicolumn{2}{c}{Macro} \\
\midrule
\multicolumn{12}{l}{\textit{XXS}} \\
\quad Transf. & 2 & 18.3/1.305 & 28.4/1.075 & 59.2/1.281 & 43.4/1.205 & 23.7/1.239 & 50.8/1.272 & 27.6/1.586 & 63.7/1.302 & 39.41 & 1.283 \\
\quad SPS & 2 & 20.2/1.227 & 29.1/1.056 & 59.4/1.256 & 44.1/1.181 & 23.9/1.214 & 50.2/1.247 & 27.9/1.577 & 68.0/1.216 & 40.35 & 1.247 \\
\quad Two-tower 3+3 & 2 & 17.5/1.326 & 28.8/1.070 & 59.2/1.276 & 43.5/1.225 & 23.8/1.251 & 51.1/1.267 & 29.1/1.588 & 63.1/1.360 & 39.53 & 1.295 \\
\quad Seq. 3+3 & 2 & 19.4/1.240 & 29.5/1.049 & 59.1/1.247 & 43.4/1.208 & 23.8/1.232 & 50.2/1.255 & 28.1/1.565 & 65.8/1.309 & 39.91 & 1.263 \\
\midrule
\multicolumn{12}{l}{\textit{XS}} \\
\quad Transf. & 2 & 21.7/1.135 & 30.6/1.012 & 60.6/1.200 & 45.3/1.131 & 24.5/1.165 & 51.7/1.219 & 30.0/1.536 & 67.7/1.277 & 41.53 & 1.209 \\
\quad SPS & 2 & 24.5/1.049 & 31.5/0.992 & 60.6/1.179 & 46.4/1.137 & 24.8/1.163 & 50.6/1.197 & 29.8/1.521 & 70.1/1.255 & 42.29 & 1.187 \\
\quad Two-tower 4+4 & 2 & 21.9/1.133 & 30.8/1.005 & 61.3/1.196 & 46.2/1.132 & 25.3/1.163 & 49.0/1.218 & 29.0/1.533 & 67.9/1.248 & 41.42 & 1.203 \\
\quad Seq. 4+4 & 2 & 24.4/1.066 & 31.6/0.988 & 61.3/1.172 & 46.6/1.126 & 25.6/1.158 & 50.4/1.199 & 30.6/1.519 & 71.0/1.210 & 42.70 & 1.180 \\
\midrule
\multicolumn{12}{l}{\textit{S}} \\
\quad Transf. & 2 & 27.8/0.929 & 35.3/0.938 & 64.7/1.116 & 48.6/1.084 & 27.7/1.106 & 51.9/1.161 & 30.9/1.468 & 72.2/1.179 & 44.89 & 1.123 \\
\quad SPS & 2 & 31.6/0.834 & 37.3/0.915 & 63.9/1.088 & 49.7/1.055 & 25.8/1.092 & 51.5/1.139 & 31.9/1.449 & 73.9/1.164 & 45.71 & 1.092 \\
\quad SPS (tied head) & 2 & 32.2/0.830 & 37.1/0.919 & 65.3/1.091 & 48.7/1.088 & 27.1/1.100 & 51.2/1.142 & 32.1/1.449 & 73.7/1.187 & 45.92 & 1.101 \\
\quad Two-tower 6+6 & 2 & 28.4/0.913 & 36.0/0.929 & 64.8/1.102 & 49.7/1.065 & 26.0/1.096 & 52.0/1.157 & 32.2/1.472 & 72.6/1.214 & 45.21 & 1.119 \\
\quad Seq. 6+6 & 2 & 30.4/0.865 & 37.3/0.923 & 65.0/1.091 & 49.3/1.073 & 26.5/1.097 & 52.1/1.146 & 30.0/1.462 & 74.3/1.177 & 45.60 & 1.104 \\
\quad Seq.\ 6+6, tied, $\langle$p$\rangle$ & 2 & 31.0/0.856 & 36.7/0.924 & 64.5/1.099 & 49.0/1.107 & 26.5/1.121 & 52.7/1.147 & 32.7/1.459 & 74.0/1.169 & 45.89 & 1.110 \\
\quad Seq.\ 6+6, $\langle$p$\rangle$ input & 1 & 30.5/0.857 & 37.2/0.918 & 64.6/1.096 & 48.9/1.097 & 26.8/1.109 & 53.4/1.145 & 31.8/1.454 & 71.8/1.218 & 45.62 & 1.112 \\
\quad Transf. 24L & 1 & 31.3/0.854 & 38.6/0.907 & 65.0/1.081 & 50.1/1.026 & 24.7/1.057 & 51.9/1.136 & 32.4/1.440 & 73.4/1.121 & 45.93 & 1.078 \\
\quad Two-tower 12+12 & 2 & 31.7/0.823 & 39.5/0.898 & 65.9/1.069 & 51.8/1.015 & 27.6/1.049 & 51.7/1.133 & 32.0/1.445 & 73.8/1.131 & 46.75 & 1.070 \\
\quad Seq. 12+12 & 2 & 33.6/0.787 & 40.9/0.888 & 66.5/1.056 & 51.8/1.012 & 27.6/1.034 & 52.9/1.123 & 33.6/1.436 & 74.0/1.136 & 47.62 & 1.059 \\
\midrule
\multicolumn{12}{l}{\textit{M}} \\
\quad Transf. & 1 & 33.4/0.776 & 42.5/0.874 & 68.2/1.034 & 56.0/0.936 & 28.8/0.983 & 52.2/1.112 & 33.8/1.407 & 79.1/0.956 & 49.24 & 1.010 \\
\quad SPS & 2 & 37.8/0.695 & 46.1/0.852 & 68.5/1.003 & 56.9/0.935 & 29.8/0.976 & 54.5/1.092 & 35.1/1.391 & 79.8/0.962 & 51.06 & 0.988 \\
\quad Two-tower 12+12 & 1 & 35.1/0.751 & 43.0/0.868 & 68.1/1.025 & 53.8/0.984 & 28.5/1.014 & 52.8/1.113 & 33.8/1.410 & 77.3/1.046 & 49.06 & 1.027 \\
\quad Seq. 12+12 & 2 & 36.2/0.724 & 44.5/0.860 & 67.9/1.026 & 55.9/0.950 & 30.0/0.988 & 53.2/1.102 & 34.8/1.402 & 78.4/1.025 & 50.10 & 1.010 \\
\midrule
\multicolumn{12}{l}{\textit{L}} \\
\quad Seq. 18+18 & 1 & 40.7/0.644 & 49.8/0.829 & 69.9/0.971 & 60.6/0.879 & 34.3/0.927 & 54.9/1.073 & 36.6/1.373 & 79.1/0.999 & 53.24 & 0.962 \\
\bottomrule
\end{tabular*}
}

\end{table}

\begin{table}[!tb]
\centering
\caption{From XXS to S, Sequential beats the Transformer and Two-tower on bits per byte in both seeds. Paired bootstrap differences with 95\% intervals: each model minus the same-size Transformer ($^{\dagger}$: Transformer 24L), then the direct pairs quoted in the text. Seed-avg: examples averaged over seeds; seed $k$: seed $k$ of both. $^{*}$: interval excludes zero.}
\label{tab:ds-paired}
{\scriptsize\setlength{\tabcolsep}{1pt}
\begin{tabular*}{\linewidth}{@{\extracolsep{\fill}}lccc@{}}
\toprule
Model & seed-avg & seed 1 & seed 2 \\
\midrule
\multicolumn{4}{@{}l}{\emph{Macro accuracy, model minus Transformer (points)}} \\
XXS SPS & +0.94$^{*}$ {\tiny[+0.43,+1.44]} & +1.03$^{*}$ {\tiny[+0.32,+1.75]} & +0.84$^{*}$ {\tiny[+0.14,+1.54]} \\
XXS Two-tower 3+3 & +0.12 {\tiny[$-$0.39,+0.64]} & +0.19 {\tiny[$-$0.53,+0.92]} & +0.05 {\tiny[$-$0.67,+0.76]} \\
XXS Seq.\ 3+3 & +0.50 {\tiny[$-$0.02,+0.99]} & +1.30$^{*}$ {\tiny[+0.60,+2.01]} & $-$0.30 {\tiny[$-$1.03,+0.39]} \\
XS SPS & +0.76$^{*}$ {\tiny[+0.24,+1.26]} & +0.87$^{*}$ {\tiny[+0.15,+1.60]} & +0.65 {\tiny[$-$0.07,+1.35]} \\
XS Two-tower 4+4 & $-$0.11 {\tiny[$-$0.64,+0.42]} & $-$0.14 {\tiny[$-$0.88,+0.60]} & $-$0.08 {\tiny[$-$0.83,+0.66]} \\
XS Seq.\ 4+4 & +1.17$^{*}$ {\tiny[+0.64,+1.69]} & +1.39$^{*}$ {\tiny[+0.66,+2.11]} & +0.95$^{*}$ {\tiny[+0.22,+1.68]} \\
S SPS & +0.82$^{*}$ {\tiny[+0.31,+1.34]} & +0.57 {\tiny[$-$0.14,+1.29]} & +1.08$^{*}$ {\tiny[+0.36,+1.80]} \\
S Two-tower 6+6 & +0.33 {\tiny[$-$0.18,+0.83]} & +0.09 {\tiny[$-$0.62,+0.80]} & +0.56 {\tiny[$-$0.14,+1.28]} \\
S Seq.\ 6+6 & +0.72$^{*}$ {\tiny[+0.19,+1.24]} & +0.39 {\tiny[$-$0.33,+1.13]} & +1.04$^{*}$ {\tiny[+0.31,+1.75]} \\
S Seq.\ 6+6 $\langle$predict$\rangle$ & +0.74$^{*}$ {\tiny[+0.10,+1.38]} & +0.39 {\tiny[$-$0.34,+1.12]} & +1.09$^{*}$ {\tiny[+0.34,+1.84]} \\
S Transf.\ 24L & +1.05$^{*}$ {\tiny[+0.43,+1.66]} & +0.70 {\tiny[+0.00,+1.42]} & +1.40$^{*}$ {\tiny[+0.67,+2.11]} \\
S Two-tower 12+12$^{\dagger}$ & +0.82$^{*}$ {\tiny[+0.20,+1.44]} & +0.76$^{*}$ {\tiny[+0.06,+1.46]} & +0.89$^{*}$ {\tiny[+0.17,+1.59]} \\
S Seq.\ 12+12$^{\dagger}$ & +1.69$^{*}$ {\tiny[+1.05,+2.33]} & +1.93$^{*}$ {\tiny[+1.19,+2.67]} & +1.45$^{*}$ {\tiny[+0.71,+2.17]} \\
M SPS & +1.82$^{*}$ {\tiny[+1.18,+2.45]} & +1.91$^{*}$ {\tiny[+1.18,+2.63]} & +1.72$^{*}$ {\tiny[+0.99,+2.43]} \\
M Two-tower 12+12 & $-$0.19 {\tiny[$-$0.90,+0.55]} & $-$0.19 {\tiny[$-$0.90,+0.55]} & -- \\
M Seq.\ 12+12 & +0.85$^{*}$ {\tiny[+0.23,+1.47]} & +0.81$^{*}$ {\tiny[+0.10,+1.55]} & +0.89$^{*}$ {\tiny[+0.17,+1.61]} \\
\midrule
\multicolumn{4}{@{}l}{\emph{Macro gold bits per byte, model minus Transformer}} \\
XXS SPS & $-$0.0364$^{*}$ {\tiny[$-$0.0386,$-$0.0342]} & $-$0.0326$^{*}$ {\tiny[$-$0.0356,$-$0.0296]} & $-$0.0402$^{*}$ {\tiny[$-$0.0431,$-$0.0371]} \\
XXS Two-tower 3+3 & +0.0122$^{*}$ {\tiny[+0.0100,+0.0144]} & +0.0002 {\tiny[$-$0.0028,+0.0031]} & +0.0243$^{*}$ {\tiny[+0.0213,+0.0272]} \\
XXS Seq.\ 3+3 & $-$0.0199$^{*}$ {\tiny[$-$0.0221,$-$0.0177]} & $-$0.0225$^{*}$ {\tiny[$-$0.0254,$-$0.0195]} & $-$0.0174$^{*}$ {\tiny[$-$0.0204,$-$0.0144]} \\
XS SPS & $-$0.0226$^{*}$ {\tiny[$-$0.0249,$-$0.0204]} & $-$0.0296$^{*}$ {\tiny[$-$0.0326,$-$0.0266]} & $-$0.0156$^{*}$ {\tiny[$-$0.0187,$-$0.0127]} \\
XS Two-tower 4+4 & $-$0.0060$^{*}$ {\tiny[$-$0.0081,$-$0.0038]} & $-$0.0032$^{*}$ {\tiny[$-$0.0062,$-$0.0001]} & $-$0.0087$^{*}$ {\tiny[$-$0.0117,$-$0.0059]} \\
XS Seq.\ 4+4 & $-$0.0296$^{*}$ {\tiny[$-$0.0319,$-$0.0272]} & $-$0.0373$^{*}$ {\tiny[$-$0.0404,$-$0.0342]} & $-$0.0218$^{*}$ {\tiny[$-$0.0249,$-$0.0187]} \\
S SPS & $-$0.0305$^{*}$ {\tiny[$-$0.0328,$-$0.0283]} & $-$0.0266$^{*}$ {\tiny[$-$0.0296,$-$0.0236]} & $-$0.0344$^{*}$ {\tiny[$-$0.0373,$-$0.0314]} \\
S Two-tower 6+6 & $-$0.0039$^{*}$ {\tiny[$-$0.0061,$-$0.0017]} & $-$0.0052$^{*}$ {\tiny[$-$0.0081,$-$0.0023]} & $-$0.0026 {\tiny[$-$0.0055,+0.0004]} \\
S Seq.\ 6+6 & $-$0.0184$^{*}$ {\tiny[$-$0.0207,$-$0.0161]} & $-$0.0215$^{*}$ {\tiny[$-$0.0244,$-$0.0185]} & $-$0.0152$^{*}$ {\tiny[$-$0.0182,$-$0.0121]} \\
S Seq.\ 6+6 $\langle$predict$\rangle$ & $-$0.0111$^{*}$ {\tiny[$-$0.0136,$-$0.0085]} & $-$0.0155$^{*}$ {\tiny[$-$0.0183,$-$0.0127]} & $-$0.0066$^{*}$ {\tiny[$-$0.0096,$-$0.0036]} \\
S Transf.\ 24L & $-$0.0448$^{*}$ {\tiny[$-$0.0472,$-$0.0422]} & $-$0.0492$^{*}$ {\tiny[$-$0.0521,$-$0.0463]} & $-$0.0403$^{*}$ {\tiny[$-$0.0432,$-$0.0374]} \\
S Two-tower 12+12$^{\dagger}$ & $-$0.0075$^{*}$ {\tiny[$-$0.0100,$-$0.0050]} & $-$0.0113$^{*}$ {\tiny[$-$0.0141,$-$0.0086]} & $-$0.0037$^{*}$ {\tiny[$-$0.0065,$-$0.0008]} \\
S Seq.\ 12+12$^{\dagger}$ & $-$0.0188$^{*}$ {\tiny[$-$0.0213,$-$0.0162]} & $-$0.0221$^{*}$ {\tiny[$-$0.0250,$-$0.0192]} & $-$0.0155$^{*}$ {\tiny[$-$0.0184,$-$0.0125]} \\
M SPS & $-$0.0215$^{*}$ {\tiny[$-$0.0241,$-$0.0190]} & $-$0.0243$^{*}$ {\tiny[$-$0.0272,$-$0.0215]} & $-$0.0186$^{*}$ {\tiny[$-$0.0215,$-$0.0157]} \\
M Two-tower 12+12 & +0.0168$^{*}$ {\tiny[+0.0139,+0.0197]} & +0.0168$^{*}$ {\tiny[+0.0139,+0.0197]} & -- \\
M Seq.\ 12+12 & +0.0001 {\tiny[$-$0.0025,+0.0026]} & $-$0.0070$^{*}$ {\tiny[$-$0.0099,$-$0.0041]} & +0.0072$^{*}$ {\tiny[+0.0041,+0.0103]} \\
\midrule
\multicolumn{4}{@{}l}{\emph{Macro accuracy, Sequential minus other models (points)}} \\
XXS Seq.\ 3+3 vs Two-tower & +0.38 {\tiny[$-$0.14,+0.88]} & +1.11$^{*}$ {\tiny[+0.39,+1.82]} & $-$0.35 {\tiny[$-$1.08,+0.36]} \\
XS Seq.\ 4+4 vs Two-tower & +1.28$^{*}$ {\tiny[+0.77,+1.79]} & +1.52$^{*}$ {\tiny[+0.81,+2.24]} & +1.03$^{*}$ {\tiny[+0.31,+1.76]} \\
S Seq.\ 6+6 vs Two-tower & +0.39 {\tiny[$-$0.13,+0.89]} & +0.30 {\tiny[$-$0.39,+0.99]} & +0.48 {\tiny[$-$0.23,+1.18]} \\
S Seq.\ 12+12 vs Two-tower & +0.87$^{*}$ {\tiny[+0.35,+1.40]} & +1.17$^{*}$ {\tiny[+0.47,+1.88]} & +0.56 {\tiny[$-$0.15,+1.28]} \\
M Seq.\ 12+12 vs Two-tower & +1.04$^{*}$ {\tiny[+0.43,+1.66]} & +1.00$^{*}$ {\tiny[+0.28,+1.73]} & +1.08$^{*}$ {\tiny[+0.37,+1.77]} \\
S Seq.\ 6+6 vs SPS & $-$0.11 {\tiny[$-$0.61,+0.39]} & $-$0.18 {\tiny[$-$0.87,+0.51]} & $-$0.04 {\tiny[$-$0.76,+0.66]} \\
M Seq.\ 12+12 vs SPS & $-$0.96$^{*}$ {\tiny[$-$1.47,$-$0.47]} & $-$1.09$^{*}$ {\tiny[$-$1.81,$-$0.38]} & $-$0.83$^{*}$ {\tiny[$-$1.53,$-$0.14]} \\
\midrule
\multicolumn{4}{@{}l}{\emph{Macro gold bits per byte, Sequential minus other models}} \\
XXS Seq.\ 3+3 vs Two-tower & $-$0.0321$^{*}$ {\tiny[$-$0.0343,$-$0.0299]} & $-$0.0226$^{*}$ {\tiny[$-$0.0257,$-$0.0196]} & $-$0.0416$^{*}$ {\tiny[$-$0.0445,$-$0.0387]} \\
XS Seq.\ 4+4 vs Two-tower & $-$0.0236$^{*}$ {\tiny[$-$0.0259,$-$0.0214]} & $-$0.0341$^{*}$ {\tiny[$-$0.0371,$-$0.0313]} & $-$0.0131$^{*}$ {\tiny[$-$0.0161,$-$0.0101]} \\
S Seq.\ 6+6 vs Two-tower & $-$0.0145$^{*}$ {\tiny[$-$0.0164,$-$0.0125]} & $-$0.0163$^{*}$ {\tiny[$-$0.0191,$-$0.0137]} & $-$0.0127$^{*}$ {\tiny[$-$0.0153,$-$0.0099]} \\
S Seq.\ 12+12 vs Two-tower & $-$0.0113$^{*}$ {\tiny[$-$0.0134,$-$0.0092]} & $-$0.0107$^{*}$ {\tiny[$-$0.0136,$-$0.0080]} & $-$0.0118$^{*}$ {\tiny[$-$0.0147,$-$0.0090]} \\
M Seq.\ 12+12 vs Two-tower & $-$0.0167$^{*}$ {\tiny[$-$0.0191,$-$0.0144]} & $-$0.0238$^{*}$ {\tiny[$-$0.0266,$-$0.0211]} & $-$0.0096$^{*}$ {\tiny[$-$0.0125,$-$0.0068]} \\
S Seq.\ 6+6 vs SPS & +0.0121$^{*}$ {\tiny[+0.0100,+0.0142]} & +0.0051$^{*}$ {\tiny[+0.0023,+0.0079]} & +0.0192$^{*}$ {\tiny[+0.0162,+0.0221]} \\
M Seq.\ 12+12 vs SPS & +0.0215$^{*}$ {\tiny[+0.0196,+0.0236]} & +0.0173$^{*}$ {\tiny[+0.0143,+0.0202]} & +0.0258$^{*}$ {\tiny[+0.0230,+0.0286]} \\
\bottomrule
\end{tabular*}
}

\end{table}

\subsection{Out-of-Domain Text}
\label{app:ood}
In the seed mean, Sequential has lower bits per byte than the Transformer and Two-tower on every corpus at every size except code and PG19 at XXS and a tie on code against Two-tower at M (Tab.~\ref{tab:ood}). In individual seeds it also trails on code in four pairings (XS versus both, S 6+6 versus Two-tower, M seed 2 versus Two-tower), where seed variance is largest. At XXS, Sequential seed 1 degrades on long code and PG19 documents ($+1.32$ bits per byte on code versus the Transformer) while seed 2 beats the Transformer; similar seed-dependent degradation at late positions also occurs in XXS SPS and Two-tower seeds and in one XS and one S SPS seed, though never in the Transformer. SPS leads Sequential on every corpus except code at M; at S, SPS leads on C4, WikiText-103, and FineWeb-Edu, while Sequential 6+6 leads on code, PG19, and arXiv.

\subsection{Retrieval within 4K Tokens}
\label{app:retrieval}
Tab.~\ref{tab:retrieval4k} (right) gives passkey and multi-key NIAH accuracy within the training context. At S, Sequential 6+6 retrieves passkeys more often than the Transformer and Two-tower 6+6 and matches SPS, while Sequential 12+12 ties Two-tower 12+12. At M all four models are close on passkey and Sequential leads on multi-key NIAH. At XXS and XS Sequential trails SPS and Two-tower on passkey, and seeds disagree by more than 0.15 in several XXS/XS cells and two S cells. Every model drops near chance once the needle lies more than roughly 2.5K tokens back, because in packed 4K windows with document masking only 10\% of tokens have at least 2K tokens of visible context. Long-range retrieval is therefore evaluated after the 32K continuation (App.~\ref{app:longctx}).

\begin{table}[!tb]
\centering
\caption{Sequential has lower out-of-domain bits per byte than the Transformer on every set at S and M. Left: bits per byte (Wiki: WikiText-103; Code: Python; FW-Edu: in-domain). Right: passkey and multi-key NIAH 5-way accuracy (chance 0.2) by length within 4K, mean over seven depths.}
\label{tab:ood}
\label{tab:retrieval4k}
{\scriptsize\setlength{\tabcolsep}{1.2pt}
\begin{tabular*}{\linewidth}{@{\extracolsep{\fill}}lcccccc@{\hspace{6pt}}cccccc@{}}
\toprule
 & \multicolumn{6}{c}{Bits per byte (lower is better)} & \multicolumn{3}{c}{Passkey} & \multicolumn{3}{c}{MK-NIAH} \\
\cmidrule(lr){2-7}\cmidrule(lr){8-10}\cmidrule(l){11-13}
Model & C4 & Wiki & Code & PG19 & arXiv & FW-Edu & 1K & 2K & 4K & 1K & 2K & 4K \\
\midrule
\multicolumn{13}{l}{\textit{XXS}} \\
\quad Transf. & 1.1996 & 1.2054 & 1.6686 & 1.4407 & 1.3955 & 1.0283 & 0.28 & 0.18 & 0.13 & 0.31 & 0.32 & 0.34 \\
\quad SPS & 1.1809 & 1.1704 & 1.7043 & 1.4401 & 1.3672 & 1.0111 & 0.47 & 0.30 & 0.17 & 0.33 & 0.33 & 0.32 \\
\quad Two-tower 3+3 & 1.1980 & 1.2097 & 1.7494 & 1.4666 & 1.3957 & 1.0269 & 0.38 & 0.23 & 0.15 & 0.32 & 0.34 & 0.36 \\
\quad Seq. 3+3 & 1.1761 & 1.1762 & 2.3307 & 1.5890 & 1.3664 & 1.0053 & 0.20 & 0.12 & 0.08 & 0.29 & 0.27 & 0.27 \\
\midrule
\multicolumn{13}{l}{\textit{XS}} \\
\quad Transf. & 1.1408 & 1.1284 & 1.5687 & 1.3816 & 1.3253 & 0.9681 & 0.46 & 0.29 & 0.19 & 0.37 & 0.32 & 0.34 \\
\quad SPS & 1.1186 & 1.0895 & 2.1305 & 1.3586 & 1.2999 & 0.9469 & 0.87 & 0.53 & 0.29 & 0.35 & 0.33 & 0.34 \\
\quad Two-tower 4+4 & 1.1337 & 1.1152 & 1.5667 & 1.3796 & 1.3122 & 0.9603 & 0.82 & 0.43 & 0.31 & 0.31 & 0.31 & 0.32 \\
\quad Seq. 4+4 & 1.1160 & 1.0876 & 1.5650 & 1.3590 & 1.2960 & 0.9437 & 0.52 & 0.31 & 0.19 & 0.50 & 0.39 & 0.37 \\
\midrule
\multicolumn{13}{l}{\textit{S}} \\
\quad Transf. & 1.0644 & 1.0201 & 1.4152 & 1.2675 & 1.2223 & 0.8907 & 0.85 & 0.58 & 0.33 & 0.46 & 0.41 & 0.34 \\
\quad SPS & 1.0416 & 0.9830 & 1.5215 & 1.2563 & 1.2115 & 0.8695 & 0.94 & 0.76 & 0.35 & 0.55 & 0.42 & 0.35 \\
\quad SPS (tied head) & 1.0439 & 0.9879 & 1.2742 & 1.2449 & 1.1941 & 0.8729 & 0.96 & 0.75 & 0.45 & 0.54 & 0.45 & 0.37 \\
\quad Two-tower 6+6 & 1.0565 & 1.0013 & 1.3693 & 1.2769 & 1.2180 & 0.8823 & 0.77 & 0.53 & 0.29 & 0.47 & 0.38 & 0.36 \\
\quad Seq. 6+6 & 1.0446 & 0.9901 & 1.3557 & 1.2528 & 1.2034 & 0.8709 & 0.91 & 0.78 & 0.44 & 0.62 & 0.57 & 0.43 \\
\quad Seq.\ 6+6, tied, $\langle$p$\rangle$ & 1.0484 & 0.9896 & 1.3213 & 1.2548 & 1.2043 & 0.8761 & 0.91 & 0.69 & 0.38 & 0.59 & 0.52 & 0.39 \\
\quad Seq.\ 6+6, $\langle$p$\rangle$ input & 1.0458 & 0.9888 & 1.3670 & 1.2593 & 1.2010 & 0.8722 & 0.98 & 0.69 & 0.41 & 0.59 & 0.52 & 0.40 \\
\quad Transf. 24L & 1.0310 & 0.9756 & 1.3162 & 1.2338 & 1.1786 & 0.8568 & 0.83 & 0.57 & 0.32 & 0.52 & 0.48 & 0.36 \\
\quad Two-tower 12+12 & 1.0237 & 0.9592 & 1.2507 & 1.2237 & 1.1697 & 0.8492 & 0.95 & 0.86 & 0.49 & 0.65 & 0.57 & 0.41 \\
\quad Seq. 12+12 & 1.0139 & 0.9444 & 1.2377 & 1.2098 & 1.1550 & 0.8396 & 0.94 & 0.88 & 0.50 & 0.70 & 0.59 & 0.43 \\
\midrule
\multicolumn{13}{l}{\textit{M}} \\
\quad Transf. & 0.9992 & 0.9307 & 1.2256 & 1.1956 & 1.1357 & 0.8243 & 0.94 & 0.91 & 0.55 & 0.61 & 0.45 & 0.39 \\
\quad SPS & 0.9761 & 0.8932 & 1.2060 & 1.1540 & 1.1124 & 0.8023 & 0.98 & 0.89 & 0.53 & 0.70 & 0.52 & 0.42 \\
\quad Two-tower 12+12 & 0.9954 & 0.9172 & 1.1940 & 1.2139 & 1.1333 & 0.8201 & 0.97 & 0.89 & 0.50 & 0.69 & 0.61 & 0.44 \\
\quad Seq. 12+12 & 0.9856 & 0.9049 & 1.1948 & 1.1812 & 1.1213 & 0.8109 & 0.99 & 0.85 & 0.52 & 0.83 & 0.75 & 0.46 \\
\midrule
\multicolumn{13}{l}{\textit{L}} \\
\quad Seq. 18+18 & 0.9528 & 0.8609 & 1.0953 & 1.1335 & 1.0744 & 0.7774 & 0.99 & 0.98 & 0.56 & 0.79 & 0.80 & 0.48 \\
\bottomrule
\end{tabular*}
}

\end{table}

\subsection{Long Context at 32K Tokens}
\label{app:longctx}
\begin{figure}[htbp]
\centering
\includegraphics[width=\linewidth]{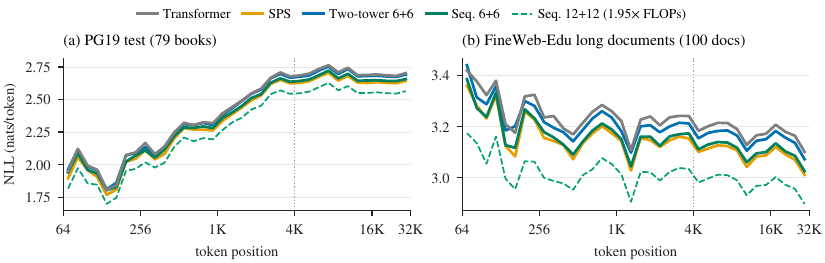}
\caption{After the 32K continuation, loss is flat or falling with position beyond 4K. Absolute NLL (nats/token) by position, in the bins of Fig.~\ref{fig:longctx}: (a) PG19, (b) long FineWeb-Edu; dotted: 4K training length.}
\label{fig:longctx-abs}
\end{figure}

Tab.~\ref{tab:longctx} reports per-position NLL and retrieval after the 32K continuation (one seed per model). On PG19 beyond 4K, Sequential 6+6 holds a steady lead over the Transformer ($-0.044$, $-0.044$, and $-0.043$ in the three long bins; 95\% document-bootstrap intervals $[-0.047,-0.041]$, $[-0.047,-0.041]$, and $[-0.045,-0.041]$), with loss flat or declining with position. SPS sits 0.015 below Sequential 6+6, and Two-tower 6+6 sits 0.009--0.012 below the Transformer. On 32K passkey retrieval, the Transformer reaches 0.28 and the 12+12 models reach 0.53--0.56, whereas Sequential 6+6 trails the Transformer at 16K and 32K; its errors occur at earlier depths, while needles near the end are retrieved reliably (0.86 at 16K, 0.94 at 32K). Multi-key NIAH ties across models at 16K--32K. Fig.~\ref{fig:longctx-abs} shows the absolute per-position NLL behind Fig.~\ref{fig:longctx}, averaged over documents and then within 36 log-spaced bins.

Before continued pretraining, every model degrades sharply past 4K. Even with the larger RoPE base, PG19 NLL at 16K--32K reaches 5.04 for the Transformer, 5.37 for Sequential 6+6, 5.46 for Two-tower 6+6, and 8.20 for SPS (whose doubled slot sequence and windowed prediction keys are tuned to the 4K geometry); with the original base, every model collapses. Zero-shot, the Transformer extrapolates best (Sequential 6+6 is 0.14--0.32 worse beyond 4K), and continued pretraining removes this gap and restores the 4K ordering.

\begin{table}[!htbp]
\centering
\scriptsize
\caption{After the 32K continuation, Sequential 6+6's loss lead is flat with position, but it finds passkeys less often than the Transformer at 16K and 32K. Top: NLL minus the Transformer's (nats/token) by position bin. Bottom: retrieval accuracy by context length (standard error $\le0.027$). One seed each.}
\label{tab:longctx}
\setlength{\tabcolsep}{2.2pt}
\begin{tabular*}{\linewidth}{@{\extracolsep{\fill}}lrrrrrrrrrr@{}}
\toprule
 & \multicolumn{4}{c}{PG19, $\Delta$ vs Transformer} & & \multicolumn{4}{c}{FineWeb-Edu long, $\Delta$ vs Transformer} & \\
\cmidrule(lr){2-5}\cmidrule(lr){7-10}
Model & 2K--4K & 4K--8K & 8K--16K & 16K--32K & & 2K--4K & 4K--8K & 8K--16K & 16K--32K & \\
\midrule
SPS & $-0.053$ & $-0.058$ & $-0.061$ & $-0.058$ & & $-0.086$ & $-0.087$ & $-0.084$ & $-0.090$ & \\
Two-tower 6+6 & $-0.008$ & $-0.012$ & $-0.009$ & $-0.009$ & & $-0.030$ & $-0.029$ & $-0.022$ & $-0.026$ & \\
Sequential 6+6 & $-0.042$ & $-0.044$ & $-0.044$ & $-0.043$ & & $-0.073$ & $-0.074$ & $-0.071$ & $-0.076$ & \\
Two-tower 12+12 & $-0.103$ & $-0.106$ & $-0.106$ & $-0.106$ & & $-0.159$ & $-0.159$ & $-0.152$ & $-0.158$ & \\
Sequential 12+12 & $-0.132$ & $-0.137$ & $-0.139$ & $-0.137$ & & $-0.209$ & $-0.203$ & $-0.199$ & $-0.205$ & \\
\midrule
 & \multicolumn{5}{c}{Passkey accuracy} & \multicolumn{5}{c}{Multi-key NIAH accuracy} \\
\cmidrule(lr){2-6}\cmidrule(lr){7-11}
Model & 2K & 4K & 8K & 16K & 32K & 2K & 4K & 8K & 16K & 32K \\
\midrule
Transformer & 0.75 & 0.73 & 0.73 & 0.55 & 0.28 & 0.40 & 0.40 & 0.37 & 0.41 & 0.31 \\
SPS & 0.85 & 0.85 & 0.81 & 0.51 & 0.13 & 0.45 & 0.48 & 0.46 & 0.48 & 0.39 \\
Two-tower 6+6 & 0.81 & 0.86 & 0.76 & 0.72 & 0.16 & 0.38 & 0.43 & 0.44 & 0.39 & 0.33 \\
Sequential 6+6 & 0.92 & 0.91 & 0.81 & 0.36 & 0.15 & 0.65 & 0.56 & 0.47 & 0.40 & 0.36 \\
Two-tower 12+12 & 0.93 & 0.95 & 0.93 & 0.81 & 0.53 & 0.54 & 0.46 & 0.53 & 0.48 & 0.43 \\
Sequential 12+12 & 0.91 & 0.94 & 0.93 & 0.91 & 0.56 & 0.73 & 0.67 & 0.57 & 0.49 & 0.37 \\
\bottomrule
\end{tabular*}
\end{table}

\section{Extended Related Work}
\label{app:related-top}
\label{app:relatedext}
The architectures below separate or reuse computation in different ways, but none varies the read level with separate weights, the one variable Sequential changes.

\paragraph{Shared computation.}
DBT \citep{liu2026decode} reads same-depth primary K/V without writing K/V of its own, and shares attention, MLP and output matrices. PHD-Transformer \citep{wu2025efficient}, PLT \citep{wu2025parallel}, ParScale \citep{chen2025parallel} and Hidden Decoding \citep{liu2026hidden} also add per-token compute through shared weights. None gives the streams separate weights, so none can isolate the read choice (AF-SPS: App.~\ref{app:afsps}).

\paragraph{Cross-layer K/V reuse.}
LCKV \citep{wu2024layer} reads top-layer K/V after warmup layers and trains through iterative parallel computation. YOCO's cross-decoder reads the final K/V of its self-decoder \citep{sun2024yoco}, and Cross-Layer Attention (CLA) shares K/V between adjacent layers \citep{brandon2024cla}. \citet{wu2025systematic} unify these schemes and find that pairing queries of all layers with K/V of upper layers works best under strong cache reduction, at extra training and prefill cost. SwiftKV \citep{qiao2024swiftkv} fills later layers' K/V from an earlier layer's output to shorten prefill, and Gemma 3n shares a middle layer's K/V with all upper layers \citep{gemma2025gemma3n}. All of these continue one residual stream rather than separating state and prediction. Sequential is closest to YOCO's cross-decoder (App.~\ref{app:yoco}). The Feedback Transformer reads a weighted mixture of past-token layers at the cost of sequential training \citep{fan2020feedback}.

\paragraph{Encoder--decoder reads.}
Machine-translation studies compare same-depth reads \citep{he2018layerwise}, learned mixtures over encoder layers \citep{bapna2018training}, lower-layer access \citep{domhan2018much}, embedding-driven fusion \citep{liu2021understanding} and encoder/decoder depth splits \citep{kasai2021deep}, all over a separate source sequence. Decoder-only versus encoder--decoder comparisons \citep{wang2022what,elfeki2025return,zhang2025encoder,ho2024block} read final state but change the whole architecture rather than only the read level. XLNet shares weights across two attention streams \citep{yang2020xlnet}. Our multi-level read (App.~\ref{app:multilevel}) is a learned layer mixture in the spirit of \citet{bapna2018training}, applied to a causal state stream.

\end{document}